\documentclass{article}

\usepackage[preprint]{nips_2018}

\usepackage[utf8]{inputenc} 
\usepackage[T1]{fontenc}    
\usepackage{hyperref}       
\usepackage{url}            
\usepackage{booktabs}       
\usepackage{amsfonts}       
\usepackage{nicefrac}       
\usepackage{microtype}      
\usepackage{stackengine}
\usepackage{wrapfig}
\usepackage{lipsum}
\usepackage{subcaption}
\usepackage[font=small,labelfont=bf]{caption}

\usepackage[mathlines,switch]{lineno} 
\usepackage{multirow}
\usepackage{multicol}
\usepackage{algorithm}
\usepackage{algorithmic}

\usepackage{amsmath}
\DeclareMathOperator*{\argmax}{arg\,max}
\DeclareMathOperator*{\argmin}{arg\,min}

\usepackage[title]{appendix}
\usepackage[export]{adjustbox}
\usepackage{enumitem}
\usepackage[parfill]{parskip}
\usepackage{xargs}
\usepackage[colorinlistoftodos,textsize=tiny]{todonotes}

\usepackage{defs}

\setlist[itemize]{noitemsep,topsep=-5pt}
\setlist[enumerate]{noitemsep,topsep=-5pt}

\newcommand{\blue}[1]{{\color{blue}{#1}}}

\newcommand{\trn}{{\textrm{trn}}}
\newcommand{\val}{{\textrm{val}}}

\makeatletter
\newcommand*{\centerfloat}{%
  \parindent \z@
  \leftskip \z@ \@plus 1fil \@minus \textwidth
  \rightskip\leftskip
  \parfillskip \z@skip}
\makeatother

\title{Bayesian Model-Agnostic Meta-Learning}

\author{
Taesup Kim$^{*\ddagger2}$, Jaesik Yoon$^{*3}$, 
\textbf{Ousmane Dia}$^1$, \textbf{Sungwoong Kim}$^{4}$,\\ \textbf{Yoshua Bengio}$^{2,5}$\textbf{,} \textbf{Sungjin Ahn}$^{\ddagger6}$\\
\\
$^1$Element AI, $^2$MILA Université de Montréal, $^3$SAP, $^4$Kakao Brain,\\$^5$CIFAR Senior Fellow, $^6$Rutgers University
\thanks{$^*$Equal contribution, Correspondence to {\tt sungjin.ahn@rutgers.edu}, $\ddagger$Work also done while working at Element AI
}
}

\renewcommand\footnotemark{}

\begin{document}

\maketitle

\begin{abstract}
Due to the inherent model uncertainty, learning to infer Bayesian posterior from a few-shot dataset is an important step towards robust meta-learning.~In this paper, we propose a novel Bayesian model-agnostic meta-learning method.~The proposed method combines efficient gradient-based meta-learning with nonparametric variational inference in a principled probabilistic framework.~Unlike previous methods, during fast adaptation, the method is capable of learning complex uncertainty structure beyond a simple Gaussian approximation, and during meta-update, a novel Bayesian mechanism prevents meta-level overfitting.~Remaining a gradient-based method, it is also the first Bayesian model-agnostic meta-learning method applicable to various tasks including reinforcement learning.~Experiment results show the accuracy and robustness of the proposed method in sinusoidal regression, image classification, active learning, and reinforcement learning.

\end{abstract}

\section{Introduction}
Two-year-old children can infer a new category from only one instance \citep{smith2017developmental}. This is presumed to be because during early learning, a human brain develops foundational structures such as the ``shape bias'' in order to learn the learning procedure \citep{landau1988importance}.
This ability, also known as \textit{learning to learn} or \textit{meta-learning} \citep{biggs1985role, bengio1990learning}, 
has recently obtained much attention in machine learning by formulating it as few-shot learning \citep{lake2015human, vinyals16matching}.
Because, initiating the learning from scratch, a neural network can hardly learn anything meaningful from such a few data points, a learning algorithm should be able to extract the statistical regularity from past tasks to enable warm-start for subsequent tasks. 

Learning a new task from a few examples inherently induces a significant amount of uncertainty.~This is apparent when we train a complex model such as a neural network using only a few examples.~It is also empirically supported by the fact that a challenge in existing few-shot learning algorithms is their tendency to overfit \citep{mishra2017meta}.~A robust meta-learning algorithm therefore must be able to systematically deal with such uncertainty in order to be applicable to critical problems such as healthcare and self-driving cars.~Bayesian inference provides a principled way to address this issue.~It brings us not only robustness to overfitting but also numerous benefits such as improved prediction accuracy by Bayesian ensembling \citep{DBLP:journals/corr/BalanRMW15}, active learning \citep{Gal2016Active}, and principled/safe exploration in reinforcement learning \citep{houthooft2016vime}.~Therefore, developing a Bayesian few-shot learning method is an important step towards robust meta-learning.  

Motivated by the above arguments, in this paper we propose a Bayesian meta-learning method, called Bayesian MAML. By introducing Bayesian methods for fast adaptation and meta-update, the proposed method learns to quickly obtain an approximate posterior of a given unseen task and thus provides the benefits of having access to uncertainty.~Being an efficient and scalable gradient-based meta-learner which encodes the meta-level statistical regularity in the initial model parameters, our method is the first Bayesian model-agnostic meta-learning method which is thus applicable to various tasks including reinforcement learning.~Combining an efficient nonparametric variational inference method with gradient-based meta-learning in a principled probabilistic framework, it can learn complex uncertainty structures while remaining simple to implement. 

The main contributions of the paper are as follows.~We propose a novel Bayesian method for meta-learning.~The proposed method is based on a novel Bayesian fast adaptation method and a new meta-update loss called the Chaser loss.~To our knowledge, the Bayesian fast adaptation is the first in meta-learning that provides flexible capability to capture the complex uncertainty curvature of the task-posterior beyond a simple Gaussian approximation. Furthermore, unlike the previous methods, the Chaser loss prevents meta-level overfitting. In experiments, we show that our method is efficient, accurate, robust, and applicable to various problems: sinusoidal regression, image classification, reinforcement learning, and active learning. 

\section{Preliminaries}
Consider a model $y=f_{\ta}(x)$ parameterized by and differentiable w.r.t.~$\ta$.~Task $\tau$ is specified by a $K$-shot dataset $\cD_\tau$ that consists of a small number of training examples, e.g., $K$ pairs $(x_k,y_k)$ per class for classification. 
We assume that tasks are sampled from a task distribution $\tau \sim p(\cT)$ such that the sampled tasks share the statistical regularity of the task distribution. A meta-learning algorithm leverages this regularity to improve the learning efficiency of subsequent tasks. The whole \textit{dataset of tasks} is divided into training/validation/test \textit{tasksets}, and the dataset of each task is further divided into task-training/task-validation/task-test \textit{datasets}.

\textbf{Model-Agnostic Meta Learning (MAML)} proposed by \cite{maml} is a gradient-based meta-learning framework.~Because it works purely by gradient-based optimization without requiring additional parameters or model modification, it is simple and generally applicable to any model as long as the gradient can be estimated.

In Algorithm~\ref{algo:maml}, we briefly review MAML. At each meta-train iteration $t$, it performs: (i) \textit{Task-Sampling}: a mini-batch $\cT_t$ of tasks is sampled from the task distribution $p(\cT)$. Each task $\tau \in \cT_t$ provides task-train data $\cD_\tau^\text{trn}$ and task-validation data $\cD_\tau^\text{val}$. (ii) \textit{Fast Adaptation} (or \textit{Inner-Update}): the parameter for each task $\tau$ in $\cT_t$ is updated by starting from the \textit{current} generic initial model $\ta_0$ and then performing $n$ gradient descent steps on the task-train loss, an operation which we denote by $\text{GD}_n(\ta_0;\cD_\tau^\trn,\al)$ with $\al$ being a step size. (iii) \textit{Meta-Update} (or \textit{Outer-Update}): the generic initial parameter $\ta_0$ is updated by gradient descent. The meta-loss is the summation of task-validation losses for all tasks in $\cT_t$, i.e., $\sum \cL(\ta_\tau;\cD_\tau^\val)$ where the summation is over all ${\tau \in \cT_t}$. 
At meta-test time, given an unseen test-task $\bar{\tau} \sim p(\cT)$, starting from the optimized initial model $\ta_0^{*}$, we obtain a model $\ta_{\bar{\tau}}$ by taking a small number of  inner-update steps using $K$-shot \textit{task-training} data $\cD_{\bar{\tau}}^{\text{trn}}$. Then, the learned model $\ta_{\bar{\tau}}$ is evaluated on the \textit{task-test} dataset $\cD_{\bar{\tau}}^{\text{tst}}$.

\textbf{Stein Variational Gradient Descent (SVGD)} \citep{liu2016stein} is a recently proposed nonparametric variational inference method. SVGD combines the strengths of MCMC and variational inference.~Unlike traditional variational inference, SVGD does not confine the family of approximate distributions within tractable parametric distributions while still remaining a simple algorithm.~Also, it converges faster than MCMC because its update rule is deterministic and leverages the gradient of the target distribution. 
Specifically, to obtain $M$ samples from target distribution $p(\ta)$, SVGD maintains $M$ instances of model parameters, called \textit{particles}. We denote the particles by $\Ta = \{\ta^m\}_{m=1}^M$. At iteration $t$, each particle $\ta_t \in \Ta_t$ is updated by the following rule:
\bea
\ta_\tpo \law \ta_t + \ep_t \phi(\ta_t) \hspace{2mm} \textrm{where}\hspace{2mm}\phi(\ta_t) = \f{1}{M}\sm{j}{M}\left[ k(\ta^j_t, \ta_t)\grad_{\ta^j_t} \log p(\ta^j_t)  + \grad_{\ta^j_t} k(\ta^j_t, \ta_t)\right],
\label{eq:svgd}
\eea 
where $\ep_t$ is step-size and $k(x,x')$ is a positive-definite kernel.~We can see that a particle consults with other particles by asking their gradients, and thereby determines its own update direction. The importance of other particles is weighted according to the kernel distance, relying more on closer particles.~The last term $\grad_{\ta^j} k(\ta^j, \ta^m)$ enforces repulsive force between particles so that they do not collapse to a point.~The resulting particles can be used to obtain the posterior predictive distribution
$p(y|x,\cD^{\tau}) = \int p(y|x,\ta) p(\ta|\cD^{\tau}) \upd \ta \approx \f{1}{M}\sum_m p(y|x,\ta^m)$
where $\ta^m \sim p(\ta|\cD^{\tau})$.  

A few properties of SVGD are particularly relevant to the proposed method: (i) when the number of particles $M$ equals 1, SVGD becomes standard gradient ascent on the objective $\log p(\ta)$, (ii) under a certain condition, an SVGD-update increasingly reduces the distance between the approximate distribution defined by the particles and the target distribution, in the sense of Kullback-Leibler (KL) divergence \citep{liu2016stein}, and finally (iii) it is straightforward to apply to reinforcement learning by using Stein Variational Policy Gradient (SVPG) \citep{liu2017stein}.

\begin{figure}[!t]
\hspace{-1mm}
\begin{minipage}{.45\textwidth}
  \begin{algorithm}[H]
    \caption{MAML}
    \label{algo:maml}
    \begin{algorithmic}
        \STATE Sample a mini-batch of tasks $\cT_t$ from $p(\cT)$
        \FOR {each task $\tau \in \cT_t$}
          \STATE $\ta_{\tau} \law \textrm{GD}_n(\ta_0; \cD_\tau^\trn, \al)$ 
        \ENDFOR
        \STATE $\ta_{0} \law \ta_{0} - \bt \grad_{\ta_{0}} \sum_{\tau \in \cT_t}   \cL(\ta_\tau;\cD_\tau^\val)$
    \end{algorithmic}
  \end{algorithm}%
\end{minipage}
\hspace{1mm}
\begin{minipage}{.52\textwidth}
  \begin{algorithm}[H]
    \caption{Bayesian Fast Adaptation}
    \begin{algorithmic}
    \STATE Sample a mini-batch of tasks $\cT_t$ from $p(\cT)$
        \FOR {each task $\tau \in \cT_t$}
        \STATE $\Ta_\tau(\Ta_0) \law \textrm{SVGD}_n(\Ta_0; \cD_\tau^\trn, \alpha)$
        \ENDFOR
    \STATE $\Ta_0 \law \Ta_0 - \bt \grad_{\Ta_0} \sum_{\tau\in\cT_t}\cL_\text{BFA}(\Ta_\tau(\Ta_0);\cD_\tau^\val)$ 
    \end{algorithmic}
    \label{algo:simple_bmaml}
  \end{algorithm}
\end{minipage}
\end{figure}

\section{Proposed Method}
\subsection{Bayesian Fast Adaptation}
Our goal is to \textit{learn to infer} by developing an efficient Bayesian gradient-based meta-learning method to efficiently obtain the task-posterior $p(\ta_{\tau} | \cD_{\tau}^\trn)$ of a novel task.~As our method is in the same class as MAML -- in the sense that it encodes the meta-knowledge in the initial model by gradient-based optimization -- we first consider the following probabilistic interpretation of MAML with one inner-update step,
\begin{align}
p(\cD_\cT^\val \mid \ta_0, \cD_\cT^\trn) = \prod_{\tau \in \cT} p(\cD_\tau^\val \mid \ta'_\tau = \ta_0 + \alpha\grad_{\ta_0} \log p(\cD_\tau^\trn \mid \ta_0)), \label{eq:pmaml}
\end{align} 
where $p(\cD_\tau^\val | \ta'_\tau) = \pd{i}{|\cD_\tau^\val|} p(y_i | x_i,\ta'_\tau)$, $\cD_\cT^\trn$ denotes all task-train sets in training taskset $\cT$, and $\cD_\cT^\val$ has the same meaning but for task-validation sets. From the above, we can see that the inner-update step of MAML amounts to obtaining task model $\ta'_\tau$ from which the likelihood of the task-validation set $\cD_\tau^\val$ is computed.~The meta-update step is then to perform maximum likelihood estimation of this model w.r.t.~the initial parameter $\ta_0$.~This probabilistic interpretation can be extended further to applying empirical Bayes to a hierarchical probabilistic model \citep{grant2018recasting} as follows:
\bea
p(\cD_\cT^\val \mid \ta_0, \cD_\cT^\trn) = \prod_{\tau \in \cT} \left(\int p(\cD_\tau^\val \mid \ta_\tau)p(\ta_\tau \mid \cD_\tau^\trn, \ta_0)\upd\ta_\tau\right). \label{eq:eb}
\eea
We see that the probabilistic MAML model in Eq.~\eqref{eq:pmaml} is a special case of Eq.~\eqref{eq:eb} that approximates the task-train posterior $p(\ta_\tau|\ta_0, \cD_\tau^\trn)$ by a point estimate $\ta'_\tau$.~That is, $p(\ta_\tau|\cD_\tau^\trn, \ta_0) = \delta_{\ta'_\tau}(\ta_\tau)$ where $\delta_y(x)=1$ if $x=y$, and 0 otherwise.~To model the task-train posterior which also becomes the prior of task-validation set, \cite{grant2018recasting} used an isotropic Gaussian distribution with a fixed variance.

``\textit{Can we use a more flexible task-train posterior than a point estimate or a simple Gaussian distribution while maintaining the efficiency of gradient-based meta-learning}?'' This is an important question because as discussed in \cite{grant2018recasting}, the task-train posterior of a  Bayesian neural network (BNN) trained with a few-shot dataset would have a significant amount of uncertainty which, according to the Bayesian central limit theorem \citep{Lecam1986, ahn2012bayesian}, cannot be well approximated by a Gaussian distribution.

Our first step for designing such an algorithm starts by noticing that SVGD performs deterministic updates and thus gradients can be backpropagated through the particles.~This means that we now maintain $M$ initial particles $\Ta_0$ and by obtaining samples from the task-train posterior $p(\ta_\tau|\cD_\tau^\trn, \Ta_0)$ using SVGD (which is now conditioned on $\Ta_0$ instead of $\ta_0$), we can optimize the following Monte Carlo approximation of Eq.~\eqref{eq:eb} by computing the gradient of the meta-loss $\log p(\cD_\cT^\val|\Ta_0, \cD_\cT^\trn)$ w.r.t. $\Ta_0$,
\bea 
p(\cD_\cT^\val \mid \Ta_0, \cD_\cT^\trn) \approx \prod_{\tau \in \cT} \left( \f{1}{M}\sm{m}{M} 
p(\cD_\tau^\val \mid \ta_\tau^m)
\right) \hspace{2mm} \textrm{where} \hspace{2mm} \ta_\tau^m \sim p(\ta_\tau \mid \cD_\tau^\trn, \Ta_0).
\eea 
Being updated by gradient descent, it hence remains an efficient meta-learning method while providing a more flexible way to capture the complex uncertainty structure of the task-train posterior than a point estimate or a simple Gaussian approximation. 

Algorithm \ref{algo:simple_bmaml} describes an implementation of the above model.~Specifically, at iteration $t$, for each task $\tau$ in a sampled mini-batch $\cT_t$, the particles initialized to $\Ta_0$ are updated for $n$ steps by applying the SVGD updater, denoted by $\textrm{SVGD}_{n}(\Ta_{0}; \cD^\trn_\tau)$ -- the target distribution (the $p(\theta^j_t)$ in Eq.~\eqref{eq:svgd} is set to the task-train posterior $p(\ta_\tau|\cD^\trn_\tau) \propto p(\cD^\trn_\tau|\ta_\tau) p(\ta_\tau)$\footnote{In our experiments, we put hyperprior on the variance of the prior (mean is set to 0). Thus, the posterior of hyperparameter is automatically learned also by SVGD, i.e., the particle vectors include the prior parameters.}. This results in task-wise particles $\Ta_\tau$ for each task $\tau \in \cT_t$. Then, for the meta-update, we can use the following meta-loss, $\log p(\cD_{\cT_t}^\val|\Ta_0, \cD_{\cT_t}^\trn)$ 
\bea 
\approx \sum_{\tau \in \cT_t}\cL_\text{BFA}(\Ta_\tau(\Ta_0);\cD_\tau^\val) \hspace{2mm}\text{where}\hspace{2mm} \cL_\text{BFA}(\Ta_\tau(\Ta_0);\cD_\tau^\val) = \log \left[\f{1}{M}\sum_{m=1}^{M} p(\cD_\tau^\val|\ta_\tau^m)\right],
\label{eq:bfa}
\eea 
Here, we use $\Ta_\tau(\Ta_0)$ to explicitly denote that $\Ta_\tau$ is a function of $\Ta_0$.
Note that, by the above model, all the initial particles in $\Ta_0$ are \textit{jointly} updated in such a way as to find the best joint-formation among them. From this optimized initial particles, the task-posterior of a new task can be obtained quickly, i.e., by taking a small number of update steps, and efficiently, i.e, with a small number of samples.~We call this Bayesian Fast Adaptation (BFA). The method can be considered a Bayesian ensemble in which, unlike non-Bayesian ensemble methods, the particles interact with each other to find the best formation representing the task-train posterior.~Because SVGD with a single particle, i.e., $M=1$, is equal to gradient ascent, Algorithm \ref{algo:simple_bmaml} reduces to MAML when $M=1$.

Although the above algorithm brings the power of Bayesian inference to fast adaptation, it can be numerically unstable due to the product of the task-validation likelihood terms. More importantly, for meta-update it is not performing Bayesian inference.~Instead, it looks for the initial prior $\Ta_0$ such that SVGD-updates lead to minimizing the empirical loss on task-validation sets. Therefore, like other meta-learning methods, the BFA model can still suffer from overfitting despite the fact that we use a flexible Bayesian inference in the inner update.~The reason is somewhat apparent.~Because we perform only a small number of inner-updates while the meta-update is based on empirical risk minimization,
the initial model $\Ta_0$ can be overfitted to the task-validation sets when we use  highly complex models like deep neural networks.
Therefore, to become a fully robust meta-learning approach, it is desired for the method to retain the uncertainty during the meta-update as well while remaining an efficient gradient-based method. 

\subsection{Bayesian Meta-Learning with Chaser Loss}
Motivated by the above observation, we propose a novel meta-loss.~For this, we start by defining the loss as the dissimilarity between approximate task-train posterior $p_\tau^n \equiv p_n(\ta_\tau | \cD_\tau^\trn; \Ta_0)$ and true task-posterior $p^\infty_\tau \equiv p(\ta_\tau | \cD_\tau^\trn \cup \cD_\tau^\val)$. Note that $p_\tau^n$ is obtained by taking $n$ fast-adaptation steps from the initial model. Assuming that we can obtain samples $\Ta_\tau^n$ and $\Ta_\tau^\infty$ respectively from these two distributions, the new meta-learning objective can be written as
\bea
\argmin_{\Ta_0} \sum_\tau d_p(p^n_\tau \;\|\; p^\infty_\tau) \approx \argmin_{\Ta_0} \sum_\tau d_s(\Ta_\tau^n(\Ta_0) \;\|\; \Ta_\tau^\infty).
\eea 
Here, $d_p(p \| q)$ is a dissimilarity between two distributions $p$ and $q$, and $d_s(s_1 \| s_2)$ a distance between two sample sets.~We then want to minimize this distance using gradient w.r.t.~$\Ta_0$.~This is to find optimized $\Ta_0$ from which the task-train posterior can be obtained quickly and closely to the true task-posterior. 
However, this is intractable because we neither have access to the true posterior $p_\tau^\infty$ nor its samples $\Ta_\tau^\infty$.

To this end, we approximate $\Ta_\tau^\infty$ by $\Ta^{n+s}_\tau$. This is done by (i) obtaining $\Ta_\tau^n$ from $p_n(\ta_\tau|\cD_\tau^\trn; \Ta_0)$ and then (ii) taking $s$ additional SVGD steps with the updated target distribution $p(\ta_\tau|\cD_\tau^\trn \cup \cD_\tau^\val)$, i.e., augmented with additional observation $\cD_\tau^\val$. Although it is valid in theory not to augment the leader with the validation set, to help fast convergence we take advantage of it like other meta-learning methods. Note that, because SVGD-updates provide increasingly better approximations of the target distribution as $s$ increases, the \textit{leader} $\Ta^{n+s}_\tau$ becomes closer to the target distribution $\Ta^\infty_\tau$ than the \textit{chaser} $\Ta^n_\tau$. This gives us the following meta-loss:
\bea
\cL_\text{BMAML}(\Ta_0) = \sum_{\tau \in \cT_t} d_s(\Ta_\tau^n \;\|\; \Ta_\tau^{n+s}) = \sum_{\tau \in \cT_t}\sum_{m=1}^{M} \| \ta_\tau^{n,m} - \ta_\tau^{n+s,m}\|_2^2.
\eea 
Here, to compute the distance between the two sample sets, we make a one-to-one mapping between the leader particles and the chaser particles and compute the Euclidean distance between the paired particles. Note that we do not back-propagate through the leader particles because we use them as targets that the chaser particles follow.~A more sophisticated method like maximum mean discrepancy \citep{borgwardt2006integrating} can also be used here. In our experiments, setting $n$ and $s$ to a small number like $n=s=1$ worked well.

Minimizing the above loss w.r.t.~$\Ta_0$ places $\Ta_0$ in a region where the chaser $\Ta^n_\tau$ can efficiently \textit{chase} the leader $\Ta^{n+s}_\tau$ in $n$ SVGD-update steps starting from $\Ta_0$. Thus, we call this meta-loss the \textit{Chaser} loss. 
Because the leader converges to the posterior distribution instead of doing empirical risk minimization, it retains a proper level of uncertainty and thus prevents from meta-level overfitting.~In Algorithm~\ref{algo:bmaml_sl}, we describe the algorithm for supervised learning. One limitation of the method is that, like other ensemble methods, it needs to maintain $M$ model instances.~Because this could sometimes be an issue when training a large model, in the Experiment section we introduce a way to share parameters among the particles. 

\begin{algorithm}[t]
\begin{algorithmic}[1]
\caption{Bayesian Meta-Learning with Chaser Loss (BMAML)}
\STATE Initialize $\Ta_0$
\FOR{$t=0,\dots$ until converge}
\STATE Sample a mini-batch of tasks $\cT_t$ from $p(\cT)$
    \FOR {each task $\tau \in \cT_t$}
    \STATE Compute chaser $\Ta_\tau^n(\Ta_0) = \textrm{SVGD}_n(\Ta_0;\cD_\tau^\trn, \alpha)$  
    \STATE Compute leader $\Ta_\tau^{n+s}(\Ta_0) = \textrm{SVGD}_s(\Ta_\tau^n(\Ta_0);\cD_\tau^\trn \cup \cD_\tau^\val, \alpha)$  
    \ENDFOR
\STATE $\Ta_0 \law \Ta_0 - \bt \grad_{\Ta_0} \sum_{\tau\in\cT_t}d_s(\Ta_\tau^n(\Ta_0) \;\|\; \textrm{stopgrad}(\Ta_\tau^{n+s}(\Ta_0)))$
\ENDFOR
\label{algo:bmaml_sl}
\end{algorithmic}
\end{algorithm}

\section{Related Works}
There have been many studies in the past that formulate meta-learning and learning-to-learn from a probabilistic modeling perspective \citep{tenenbaum1999bayesian, fe2003bayesian, lawrence2004learning, daume2009bayesian}.~Since then, 
the remarkable advances in deep neural networks \citep{krizhevsky2012imagenet, goodfellow2016deep} and the introduction of new few-shot learning datasets \citep{lake2015human, Sachin2017}, have rekindled the interest in this problem from the perspective of deep networks for few-shot learning \citep{santoro16meta, vinyals16matching, snell2017prototypical, duan2016rl, maml, mishra2017meta}.~Among these, \cite{maml} proposed MAML that formulates meta-learning as gradient-based optimization. 

\cite{grant2018recasting} reinterpreted MAML as a hierarchical Bayesian model, and proposed a way to perform an implicit posterior inference.~However, unlike our proposed model, the posterior on validation set is approximated by local Laplace approximation and used a relatively complex $2^{\text{nd}}$-order optimization using K-FAC \citep{martens2015optimizing}. The fast adaptation is also approximated by a simple isotropic Gaussian with fixed variance.~As pointed by \cite{grant2018recasting}, this approximation would not work well for skewed distributions, which is likely to be the case of BNNs trained on a few-shot dataset.~The authors also pointed that their method is limited in that the predictive distribution over new data-points is approximated by a point estimate. Our method resolves these limitations.~Although it can be expensive when training many large networks, we mitigate this cost by parameter sharing among the particles.~In addition, \cite{bauer2017discriminative} also proposed Gaussian approximation of the task-posterior and a scheme of splitting the feature network and the classifier which is similar to what we used for the image classification task.~\cite{lacoste2017deep} proposed learning a distribution of stochastic input noise while fixing the BNN model parameter.

\section{Experiments}
We evaluated our proposed model (BMAML) in various few-shot learning tasks: sinusoidal regression, image classification, active learning, and reinforcement learning.~Because our method is a Bayesian ensemble, as a baseline model we used an ensemble of independent MAML models (EMAML) from which we can easily recover regular MAML by setting the number of particles to $1$. In all our experiments, we configured BMAML and EMAML to have the same network architecture and used the RBF kernel.~The experiments are designed in such a way to see the effects of uncertainty in various ways such as accuracy, robustness, and efficient exploration. 

\textbf{Regression:}~The population of the tasks is defined by a sinusoidal function $y = A\sin(wx + b) + \epsilon$ which is parameterized by amplitude $A$, frequency $w$, and phase $b$, and observation noise $\ep$. To sample a task, we sample the parameters uniformly randomly $A \in [0.1, 5.0]$, $b \in [0.0, 2\pi]$, $w \in [0.5, 2.0]$ and add observation noise from $\ep \sim \cN(0, (0.01A)^2)$. The $K$-shot dataset is obtained by sampling $x$ from $[-5.0, 5.0]$ and then by computing its corresponding $y$ with noise $\ep$. Note that, because of the highly varying frequency and observation noise, this is a more challenging setting containing more uncertainty than the setting used in \cite{maml}. For the regression model, we used a neural network with 3 layers each of which consists of 40 hidden units. 

\begin{figure}[t]
\centering
\hspace*{-0.8cm}
\includegraphics[width=1.08\textwidth]{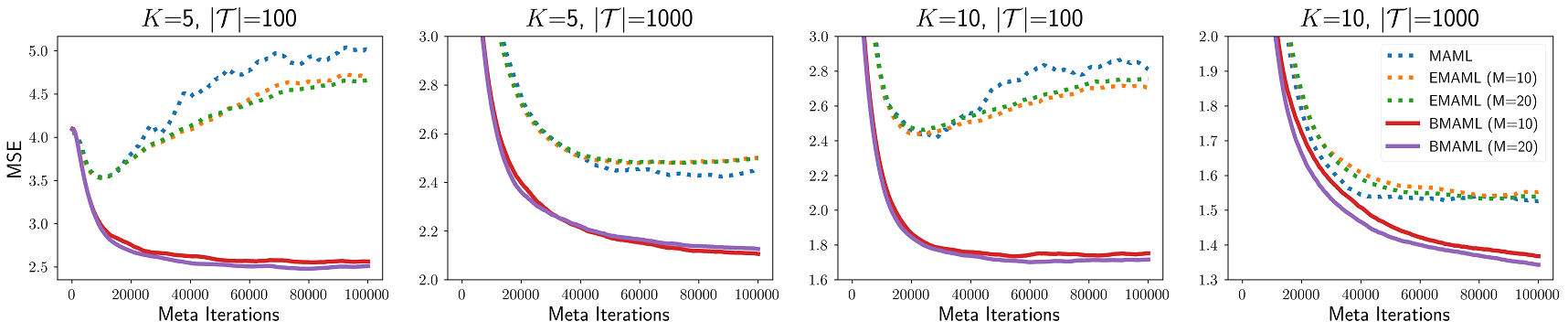}
\vspace*{-3mm}
\caption{Sinusoidal regression experimental results (meta-testing performance) by varying the number of examples~$(K\text{-shot})$ given for each task and the number of tasks $|\cT|$ used for meta-training.}
\label{fig:regress}
\end{figure}

In Fig.~\ref{fig:regress}, we show the mean squared error (MSE) performance on the test tasks. To see the effect of the degree of uncertainty, we controlled the number of training tasks $|\cT|$ to 100 and 1000, and the number of observation shots $K$ to 5 and 10. The lower number of training tasks and observation shots is expected to induce a larger degree of uncertainty.~We observe, as we claimed, that both MAML (which is EMAML with $M=1$) and EMAML overfit severely in the settings with high uncertainty although EMAML with multiple particles seems to be slightly better than MAML. BMAML with the same number of particles provides significantly better robustness and accuracy for all settings. Also, having more particles tends to improve further.

\textbf{Classification:} To evaluate the proposed method on a more complex model, we test the performance on the \textit{mini}Imagenet classification task \citep{vinyals16matching} involving task adaptation of 5-way classification with 1 shot. The dataset consists of 60,000 color images of 84$\times$84 dimension. The images consist of total 100 classes and each of the classes contains 600 examples.~The entire classes are split into 64, 12, and 24 classes for meta-train, meta-validation, and meta-test, respectively. We generated the tasks following the same procedure as in \cite{maml}.

In order to reduce the space and time complexity of the ensemble models (i.e., BMAML and EMAML) in this large network setting, we used the following parameter sharing scheme among the particles, similarly to \cite{bauer2017discriminative}.~We split the network architecture into the feature extractor layers and the classifier.~The feature extractor is a convolutional network with 5 hidden layers with $64$ filters. The classifier is a single-layer fully-connected network with softmax output.~The output of the feature extractor which has 256 dimensions is input to the classifier. We share the feature extractor across all the particles while each particle has its own classifier. Therefore, the space complexity of the network is $\cO(|\ta_\text{feature}| + M |\ta_\text{classifier}|)$. Both the classifier and feature extractor are updated during meta-update, but for inner-update only the classifier is updated. The baseline models are updated in the same manner. We describe more details of the setting in Appendix \ref{appendix:class}. 

We can see from Fig.~\ref{fig:classification} (a) that for both $M=5$ and $M=10$ BMAML provides more accurate predictions than EMAML. However, the performance of both BMAML and EMAML with 10 particles is slightly lower than having 5 particles\footnote{We found a similar instability in the relationship between the number of particles and the prediction accuracy from the original implementation by the authors of the SVGD paper.}.
Because a similar instability is also observed in the SVPG paper \citep{liu2016stein}, we presume that one possible reason is the instability of SVGD such as sensitivity to kernel function parameters.~To increase the inherent uncertainty further, in Fig.~\ref{fig:classification} (b), we reduced the number of training tasks $|\cT|$ from 800K to 10K. We see that BMAML provides robust predictions even for such a small number of training tasks while EMAML overfits easily.

\begin{figure}[t]
\centering
\begin{subfigure}[b]{0.390\textwidth}
    \centering
    \includegraphics[width=\textwidth]{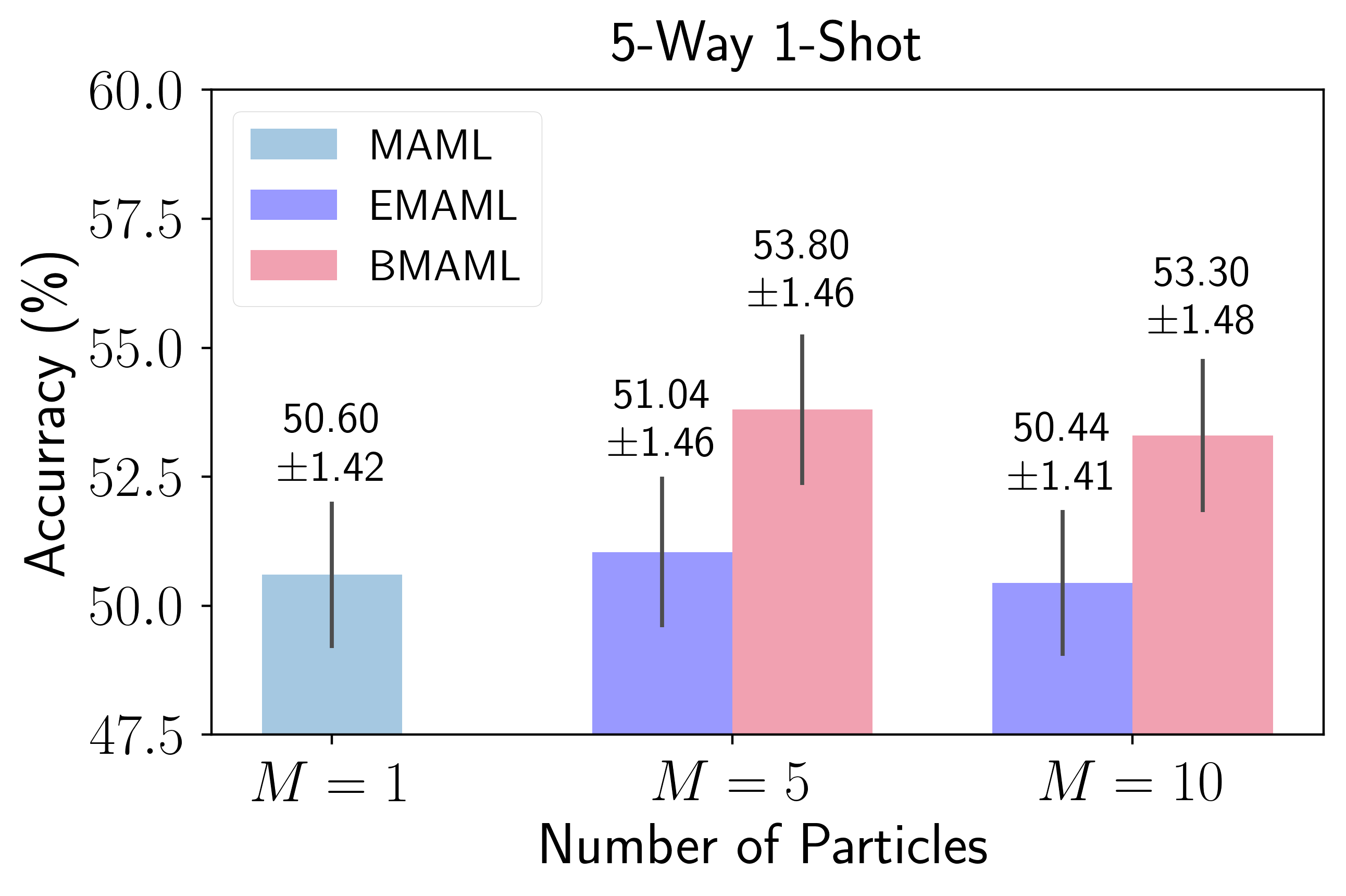}
    \caption{}
\end{subfigure}\hfill
\begin{subfigure}[b]{0.353\textwidth}
    \centering
    \includegraphics[width=\textwidth]{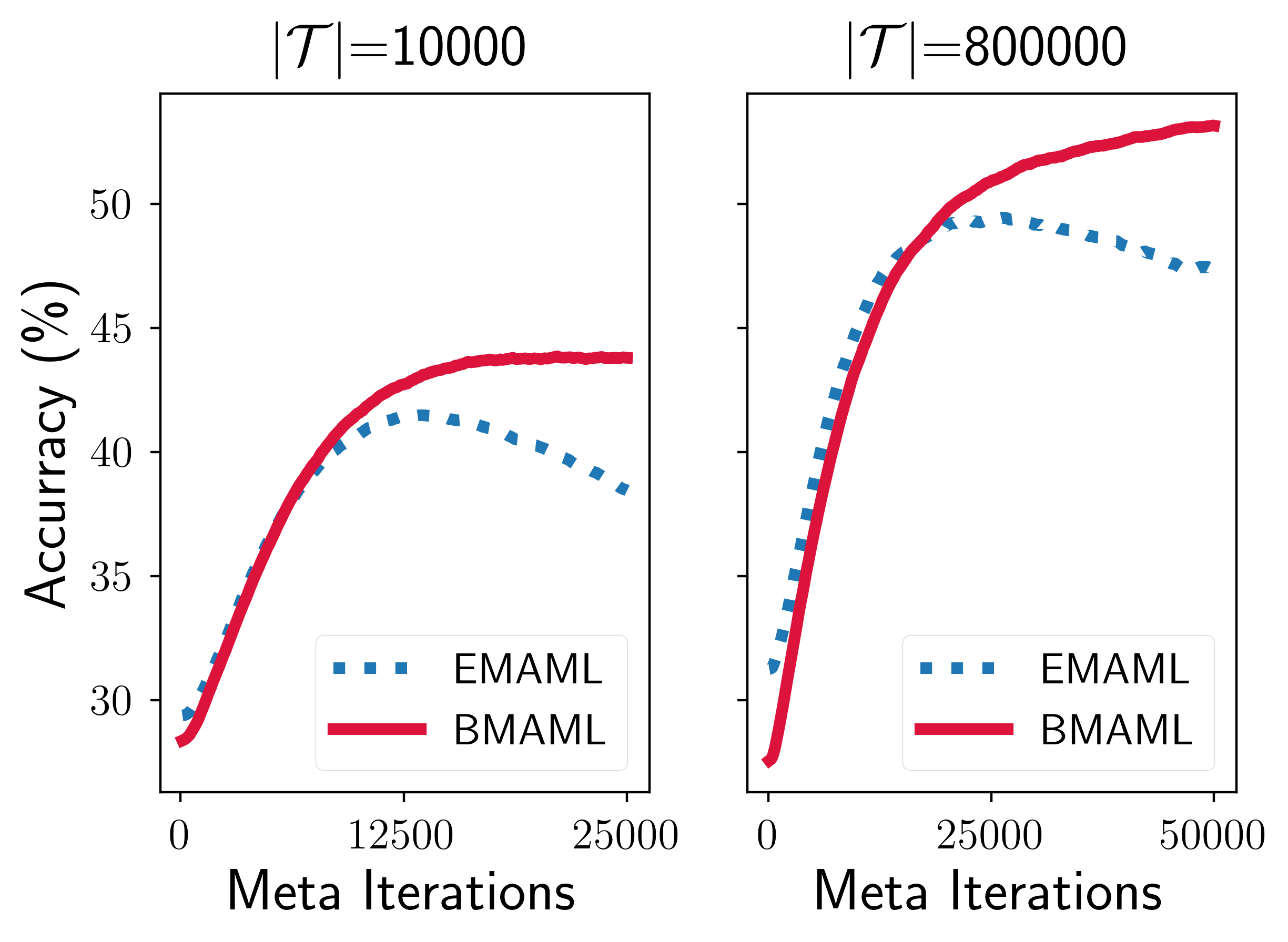}
    \caption{}
\end{subfigure}\hfill
\begin{subfigure}[b]{0.247\textwidth}
    \centering
    \includegraphics[width=\textwidth]{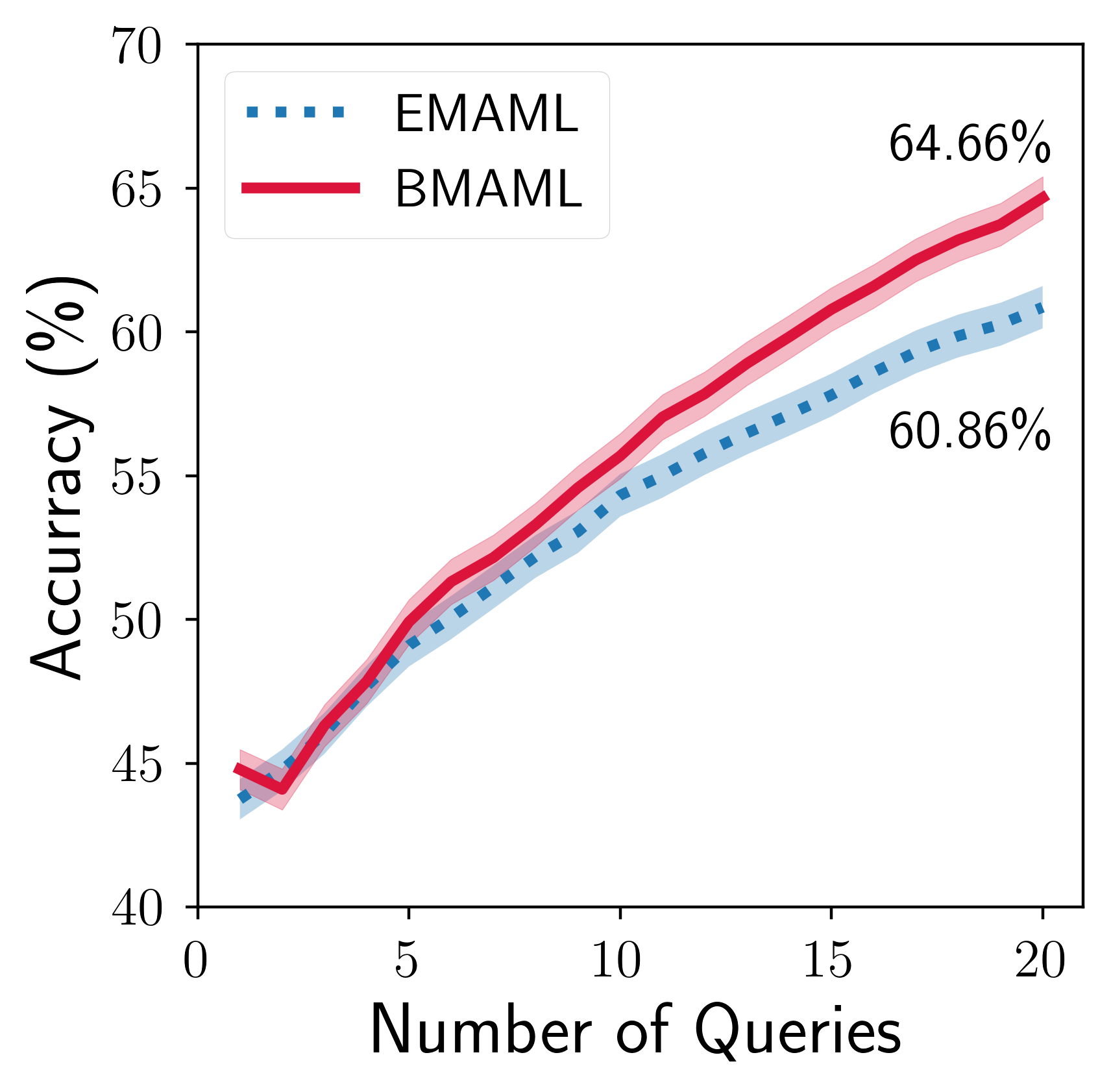}
    \caption{}
\end{subfigure}
\vspace*{-2mm}
\caption{Experimental results in \textit{mini}Imagenet dataset: (a) few-shot image classification using different number of particles, (b) using different number of tasks for meta-training, and (c) active learning setting.}
\label{fig:classification}
\end{figure}

\textbf{Active Learning:}
In addition to the ensembled prediction accuracy, we can also evaluate the effectiveness of the measured uncertainty by applying it to active learning.~To demonstrate, we use the \textit{mini}Imagenet classification task.~To do this, given an unseen task $\tau$ at test time, we first run a fast adaptation from the meta-trained initial particles $\Ta_0^*$ to obtain $\Ta_\tau$ of the task-train posterior $p(\ta_\tau|\cD_\tau; \Ta_0^*)$. For this we used 5-way 1-shot labeled dataset.~Then, from a pool of unlabeled data $\cX_\tau = \{x_1, \dots, x_{20}\}$, we choose an item $x^*$ that has the maximum predictive entropy $\argmax_{x\in \cX_{\tau}} \eH[y|x, D_\tau]=-\sum_{y'}p(y'|x,D_\tau)\log p(y'|x, D_\tau)$. The chosen item $x^*$ is then removed from $\cX_\tau$ and added to $\cD_\tau$ along with its label.~We repeat this process until we consume all the data in $\cX_\tau$. We set $M$ to $5$. As we can see from Fig.~\ref{fig:classification} (c), active learning using the Bayesian fast adaptation provides consistently better results than EMAML. Particularly, the performance gap increases as more examples are added. This shows that the examples picked by BMAML so as to reduce the uncertainty, provides proper discriminative information by capturing a reasonable approximation of the task-posterior.~We presume that the performance degradation observed in the early stage might be due to the class imbalance induced by choosing examples without considering the class balance.

\textbf{Reinforcement Learning:} 
SVPG is a simple way to apply SVGD to policy optimization.~\cite{liu2017stein} showed that the maximum entropy policy optimization can be recast to Bayesian inference.~In this framework, the particle update rule (a particle is now parameters of a policy) is simply to replace the target distribution $\log p(\ta)$ in Eq.~\eqref{eq:svgd} with the objective of the maximum entropy policy optimization, 
i.e., $\eE_{q(\ta)}[J(\ta)] + \eta\eH[q]]$ 
where $q(\ta)$ is a distribution of policies, $J(\ta)$ is the expected return of policy $\ta$, and $\eta$ is a parameter for exploration control.~Deploying multiple agents (particles) with a principled Bayesian exploration mechanism, SVPG encourages generating diverse policy behaviours while being easy to parallelize. 

We test and compare the models on the same MuJoCo continuous control tasks \citep{todorov2012mujoco} as are used in \cite{maml}. In the goal velocity task, the agent receives higher rewards as its current velocity approaches the goal velocity of the task. In the goal direction task, the reward is the magnitude of the velocity in either the forward or backward direction. We tested these tasks for two simulated robots, the ant and the cheetah. The goal velocity is sampled uniformly at random from $[0.0, 2.0]$ for the cheetah and from $[0.0,3.0]$ for the ant. As the goal velocity and the goal direction change per task, a meta learner is required to learn a given unseen task after trying $K$ episodes. We implemented the policy network with two hidden-layers each with 100 ReLU units. We tested the number of particles for $M \in \{1, 5, 10\}$ with $M=1$ only for non-ensembled MAML. We describe more details of the experiment setting in Appendix \ref{rl_locomotion}.

For meta-update, MAML uses TRPO \citep{schulman2015trust} which is designed with a special purpose to apply for reinforcement learning and uses a rather expensive $2^\text{nd}$-order optimization. However, the meta-update by the chaser loss is general-purpose and based on $1^\text{st}$-order optimization\footnote{When considering the inner update together, TRPO, Chaser and Reptile are $3^\text{rd}$/$2^\text{nd}$/$1^\text{st}$-order, respectively.}. Thus, for a fair comparison, we consider the following two experiment designs.~First, in order to evaluate the performance of the inner updates using Bayesian fast adaptation, we compare the standard MAML, which uses vanilla policy gradient (REINFORCE, \cite{williams1992simple}) for inner-updates and TRPO for meta-updates, with the Bayesian fast adaptation with TRPO meta-update. We label the former as VPG-TRPO and the later as SVPG-TRPO. Second, we compare SVPG-Chaser with VPG-Reptile. Because, similarly to the chaser loss, Reptile \citep{reptile} performs $1^\text{st}$-order gradient optimization based on the distance in the model parameter space, this provides us a fair baseline to evaluate the chaser loss in RL. The VPG-TRPO and VPG-Reptile are implemented with independent multiple agents.~We tested the comparing methods for $M=[1,5,10]$. More details of the experimental setting is provided in Appendix \ref{rl_setting}.

\begin{figure}[t]
\centering
\hspace*{-0.8cm}
\includegraphics[width=1.08\textwidth]{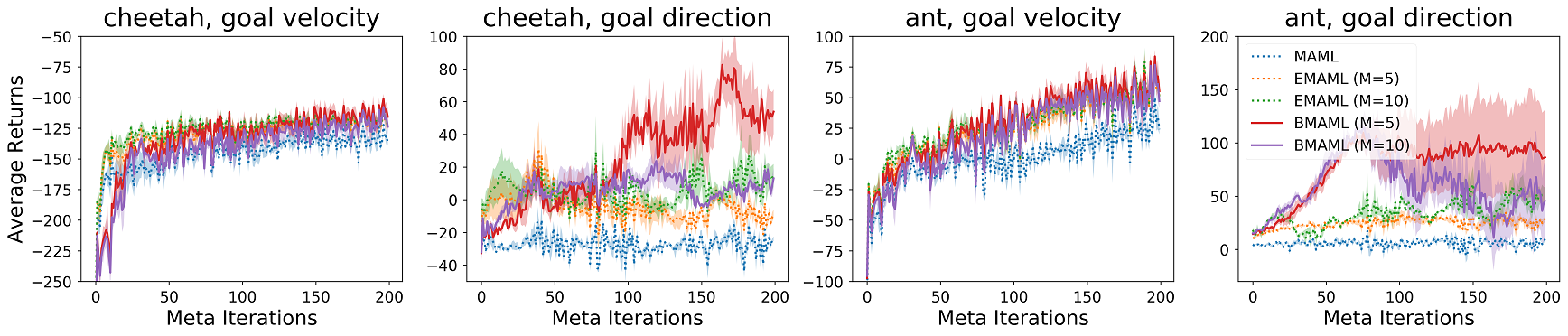}
\vspace*{-3mm}
\caption{Locomotion comparison results of SVPG-Chaser and VPG-Reptile}
\label{fig:bmaml_rl_chaser_vs_reptile_plot}
\end{figure}

As shown in Fig.~\ref{fig:bmaml_rl_trpo_vs_trpo_plot} and Fig.~\ref{fig:bmaml_rl_chaser_vs_reptile_plot}, we can see that overall, BMAML shows superior performance to EMAML. In particular, BMAML performs significantly and consistently better than EMAML in the case of using TRPO meta-updater. In addition, we can see that BMAML performs much better than EMAML for the goal direction tasks. We presume that this is because in the goal direction task, there is no goal velocity and thus a higher reward can always be obtained by searching for a better policy. This, therefore, demonstrates that BMAML can learn a better exploration policy than EMAML. In contrast, in the goal velocity task, exploration becomes less effective because it is not desired once a policy reaches the given goal velocity. This thus explains the results on the goal velocity task in which BMAML provides slightly better performance than EMAML. For some experiments, we also see that having more particles do not necessarily provides further improvements. As in the case of classification, we hypothesize that one of the reasons could be due to the instability of SVGD.~In Appendix \ref{rl_additional_res}, we also provide the results on 2D Navigation task, where we observe similar superiority of BMAML to EMAML.   

\blue{
}

\begin{figure}[t]
\centering
\hspace*{-0.8cm}
\includegraphics[width=1.08\textwidth]{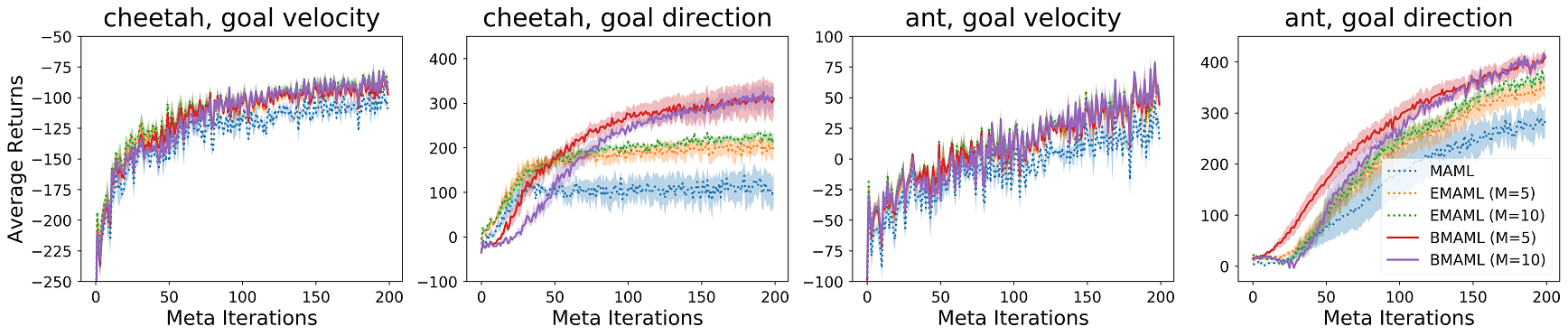}
\caption{Locomotion comparison results of SVPG-TRPO and VPG-TRPO}
\label{fig:bmaml_rl_trpo_vs_trpo_plot}
\end{figure}

\section{Discussions}
In this section, we discuss some of the issues underlying the design of the proposed method.

\textit{BMAML is tied to SVGD?} In principle, it could actually be more generally applicable to any inference algorithm that can provide \textit{differentiable samples}. Gradient-based MCMC methods like HMC \citep{neal2011mcmc} or SGLD \citep{welling2011bayesian} are such methods.~We however chose SVGD specifically for BMAML because jointly updating the particles altogether is more efficient for capturing the distribution quickly by a small number of update steps. In contrast, MCMC would require to wait for much more iterations until the chain mixes enough and a long backpropagation steps through the chain.  

\textit{Parameter space v.s.~prediction space?}~ We defined the chaser loss by the distance in the model parameter space although it is also possible to define it in the prediction distance, i.e., by prediction error. We chose the parameter space because (1) we can save computation for the forward-pass for predictions, and (2) it empirically showed better performance for RL and similar performance for other tasks.~The advantages of working in the parameter space is also discussed in \cite{reptile}. 

\textit{Do the small number of SVGD steps converge to the posterior?} In our small-data-big-network setting, a large area of a true task-posterior will be meaningless for other tasks.~Thus, it is not desired to fully capture the task-posterior but instead we need to find an area which will be broadly useful for many tasks.~This is the goal of hierarchical Bayes which our method approximate by finding such area and putting $\Ta_0$ there. In theory, the task-posterior can be fully captured with infinite number of particles and update-steps, and thus dilute the initialization effect.~In practice, the full coverage would, however, not be achievable (and not desired) because SVGD or MCMC would have difficulties in covering all areas of the complex multimodal task-posterior like that of a neural network. 

\section{Conclusion}
Motivated by the hierarchical probabilistic modeling perspective to gradient-based meta-learning, we proposed a Bayesian gradient-based meta learning method.~To do this, we combined the Stein Variational Gradient Descent with gradient-based meta learning in a probabilistic framework, and proposed the Bayesian Fast Adaptation and the Chaser loss for meta-update. As it remains a model-agnostic model, in experiments, we evaluated the method in various types of learning tasks including supervised learning, active learning, and reinforcement learning, and showed its superior performance in prediction accuracy, robustness to overfitting, and efficient exploration. 

As a Bayesian ensemble method, along with its advantages, the proposed method also inherits the generic shortcomings of ensemble methods, particularly the space/time complexity proportional to the number of particles.~Although we showed that our parameter sharing scheme is effective to mitigate this issue, it would still be interesting to improve the efficiency further in this direction. In addition, because the performance of SVGD can be sensitive to the parameters of the kernel function, incorporating the fast-adaptation of the kernel parameter into a part of meta-learning would also be an interesting future direction.

\subsubsection*{Acknowledgments}
JY thanks SAP and Kakao Brain for their support. TK thanks NSERC, MILA and Kakao Brain for their support. YB thanks CIFAR, NSERC, IBM, Google, Facebook and Microsoft for their support.~SA, Element AI Fellow, thanks Nicolas Chapados, Pedro Oliveira Pinheiro, Alexandre Lacoste, Negar Rostamzadeh for helpful discussions and feedback. 

\bibliography{bib}

\begin{thebibliography}{37}
\providecommand{\natexlab}[1]{#1}
\providecommand{\url}[1]{\texttt{#1}}
\expandafter\ifx\csname urlstyle\endcsname\relax
  \providecommand{\doi}[1]{doi: #1}\else
  \providecommand{\doi}{doi: \begingroup \urlstyle{rm}\Url}\fi

\bibitem[Ahn et~al.(2012)Ahn, Korattikara, and Welling]{ahn2012bayesian}
Sungjin Ahn, Anoop Korattikara, and Max Welling.
\newblock Bayesian posterior sampling via stochastic gradient fisher scoring.
\newblock \emph{arXiv preprint arXiv:1206.6380}, 2012.

\bibitem[Balan et~al.(2015)Balan, Rathod, Murphy, and
  Welling]{DBLP:journals/corr/BalanRMW15}
Anoop~Korattikara Balan, Vivek Rathod, Kevin Murphy, and Max Welling.
\newblock Bayesian dark knowledge.
\newblock \emph{CoRR}, abs/1506.04416, 2015.
\newblock URL \url{http://arxiv.org/abs/1506.04416}.

\bibitem[Bauer et~al.(2017)Bauer, Rojas-Carulla, {\'S}wi{\k{a}}tkowski,
  Sch{\"o}lkopf, and Turner]{bauer2017discriminative}
Matthias Bauer, Mateo Rojas-Carulla, Jakub~Bart{\l}omiej {\'S}wi{\k{a}}tkowski,
  Bernhard Sch{\"o}lkopf, and Richard~E Turner.
\newblock Discriminative k-shot learning using probabilistic models.
\newblock \emph{arXiv preprint arXiv:1706.00326}, 2017.

\bibitem[Bengio et~al.(1990)Bengio, Bengio, and Cloutier]{bengio1990learning}
Yoshua Bengio, Samy Bengio, and Jocelyn Cloutier.
\newblock \emph{Learning a synaptic learning rule}.
\newblock Universit{\'e} de Montr{\'e}al, D{\'e}partement d'informatique et de
  recherche op{\'e}rationnelle, 1990.

\bibitem[Biggs(1985)]{biggs1985role}
John~B Biggs.
\newblock The role of metalearning in study processes.
\newblock \emph{British journal of educational psychology}, 55\penalty0
  (3):\penalty0 185--212, 1985.

\bibitem[Borgwardt et~al.(2006)Borgwardt, Gretton, Rasch, Kriegel,
  Sch{\"o}lkopf, and Smola]{borgwardt2006integrating}
Karsten~M Borgwardt, Arthur Gretton, Malte~J Rasch, Hans-Peter Kriegel,
  Bernhard Sch{\"o}lkopf, and Alex~J Smola.
\newblock Integrating structured biological data by kernel maximum mean
  discrepancy.
\newblock \emph{Bioinformatics}, 22\penalty0 (14):\penalty0 e49--e57, 2006.

\bibitem[Daum{\'e}~III(2009)]{daume2009bayesian}
Hal Daum{\'e}~III.
\newblock Bayesian multitask learning with latent hierarchies.
\newblock In \emph{Proceedings of the Twenty-Fifth Conference on Uncertainty in
  Artificial Intelligence}, pp.\  135--142. AUAI Press, 2009.

\bibitem[Duan et~al.(2016)Duan, Schulman, Chen, Bartlett, Sutskever, and
  Abbeel]{duan2016rl}
Yan Duan, John Schulman, Xi~Chen, Peter~L Bartlett, Ilya Sutskever, and Pieter
  Abbeel.
\newblock Rl2: Fast reinforcement learning via slow reinforcement learning.
\newblock \emph{arXiv preprint arXiv:1611.02779}, 2016.

\bibitem[Fe-Fei et~al.(2003)]{fe2003bayesian}
Li~Fe-Fei et~al.
\newblock A bayesian approach to unsupervised one-shot learning of object
  categories.
\newblock In \emph{Computer Vision, 2003. Proceedings. Ninth IEEE International
  Conference on}, pp.\  1134--1141. IEEE, 2003.

\bibitem[Finn et~al.(2017)Finn, Abbeel, and Levine]{maml}
Chelsea Finn, Pieter Abbeel, and Sergey Levine.
\newblock Model-agnostic meta-learning for fast adaptation of deep networks.
\newblock \emph{arXiv preprint arXiv:1703.03400}, 2017.

\bibitem[Gal et~al.(2016)Gal, Islam, and Ghahramani]{Gal2016Active}
Yarin Gal, Riashat Islam, and Zoubin Ghahramani.
\newblock Deep {B}ayesian active learning with image data.
\newblock In \emph{Bayesian Deep Learning workshop, NIPS}, 2016.

\bibitem[Goodfellow et~al.(2016)Goodfellow, Bengio, Courville, and
  Bengio]{goodfellow2016deep}
Ian Goodfellow, Yoshua Bengio, Aaron Courville, and Yoshua Bengio.
\newblock \emph{Deep learning}, volume~1.
\newblock MIT press Cambridge, 2016.

\bibitem[Grant et~al.(2018)Grant, Finn, Levine, Darrell, and
  Griffiths]{grant2018recasting}
Erin Grant, Chelsea Finn, Sergey Levine, Trevor Darrell, and Thomas Griffiths.
\newblock Recasting gradient-based meta-learning as hierarchical bayes.
\newblock \emph{arXiv preprint arXiv:1801.08930}, 2018.

\bibitem[Houthooft et~al.(2016)Houthooft, Chen, Duan, Schulman, De~Turck, and
  Abbeel]{houthooft2016vime}
Rein Houthooft, Xi~Chen, Yan Duan, John Schulman, Filip De~Turck, and Pieter
  Abbeel.
\newblock Vime: Variational information maximizing exploration.
\newblock In \emph{Advances in Neural Information Processing Systems}, pp.\
  1109--1117, 2016.

\bibitem[Kingma \& Ba(2014)Kingma and Ba]{kingma2014adam}
Diederik~P Kingma and Jimmy Ba.
\newblock Adam: A method for stochastic optimization.
\newblock \emph{arXiv preprint arXiv:1412.6980}, 2014.

\bibitem[Krizhevsky et~al.(2012)Krizhevsky, Sutskever, and
  Hinton]{krizhevsky2012imagenet}
Alex Krizhevsky, Ilya Sutskever, and Geoffrey~E Hinton.
\newblock Imagenet classification with deep convolutional neural networks.
\newblock In \emph{Advances in neural information processing systems}, pp.\
  1097--1105, 2012.

\bibitem[Lacoste et~al.(2017)Lacoste, Boquet, Rostamzadeh, Oreshki, Chung, and
  Krueger]{lacoste2017deep}
Alexandre Lacoste, Thomas Boquet, Negar Rostamzadeh, Boris Oreshki, Wonchang
  Chung, and David Krueger.
\newblock Deep prior.
\newblock \emph{arXiv preprint arXiv:1712.05016}, 2017.

\bibitem[Lake et~al.(2015)Lake, Salakhutdinov, and Tenenbaum]{lake2015human}
Brenden~M Lake, Ruslan Salakhutdinov, and Joshua~B Tenenbaum.
\newblock Human-level concept learning through probabilistic program induction.
\newblock \emph{Science}, 350\penalty0 (6266):\penalty0 1332--1338, 2015.

\bibitem[Landau et~al.(1988)Landau, Smith, and Jones]{landau1988importance}
Barbara Landau, Linda~B Smith, and Susan~S Jones.
\newblock The importance of shape in early lexical learning.
\newblock \emph{Cognitive development}, 3\penalty0 (3):\penalty0 299--321,
  1988.

\bibitem[Lawrence \& Platt(2004)Lawrence and Platt]{lawrence2004learning}
Neil~D Lawrence and John~C Platt.
\newblock Learning to learn with the informative vector machine.
\newblock In \emph{Proceedings of the twenty-first international conference on
  Machine learning}, pp.\ ~65. ACM, 2004.

\bibitem[Le~Cam(1986)]{Lecam1986}
L.M. Le~Cam.
\newblock \emph{Asymptotic methods in statistical decision theory}.
\newblock Springer, 1986.

\bibitem[Liu \& Wang(2016)Liu and Wang]{liu2016stein}
Qiang Liu and Dilin Wang.
\newblock Stein variational gradient descent: A general purpose bayesian
  inference algorithm.
\newblock In \emph{Advances In Neural Information Processing Systems}, pp.\
  2378--2386, 2016.

\bibitem[Liu et~al.(2017)Liu, Ramachandran, Liu, and Peng]{liu2017stein}
Yang Liu, Prajit Ramachandran, Qiang Liu, and Jian Peng.
\newblock Stein variational policy gradient.
\newblock \emph{arXiv preprint arXiv:1704.02399}, 2017.

\bibitem[Martens \& Grosse(2015)Martens and Grosse]{martens2015optimizing}
James Martens and Roger Grosse.
\newblock Optimizing neural networks with kronecker-factored approximate
  curvature.
\newblock In \emph{International conference on machine learning}, pp.\
  2408--2417, 2015.

\bibitem[Mishra et~al.(2017)Mishra, Rohaninejad, Chen, and
  Abbeel]{mishra2017meta}
Nikhil Mishra, Mostafa Rohaninejad, Xi~Chen, and Pieter Abbeel.
\newblock Meta-learning with temporal convolutions.
\newblock \emph{arXiv preprint arXiv:1707.03141}, 2017.

\bibitem[Neal et~al.(2011)]{neal2011mcmc}
Radford~M Neal et~al.
\newblock Mcmc using hamiltonian dynamics.
\newblock \emph{Handbook of Markov Chain Monte Carlo}, 2\penalty0
  (11):\penalty0 2, 2011.

\bibitem[{Nichol} et~al.(2018){Nichol}, {Achiam}, and {Schulman}]{reptile}
A.~{Nichol}, J.~{Achiam}, and J.~{Schulman}.
\newblock {On First-Order Meta-Learning Algorithms}.
\newblock \emph{ArXiv e-prints}, March 2018.

\bibitem[Ravi \& Larochelle(2017)Ravi and Larochelle]{Sachin2017}
Sachin Ravi and Hugo Larochelle.
\newblock Optimization as a model for few-shot learning.
\newblock In \emph{In International Conference on Learning Representations
  (ICLR)}, 2017.

\bibitem[Santoro et~al.(2016)Santoro, Bartunov, Botvinick, Wierstra, and
  Lillicrap]{santoro16meta}
Adam Santoro, Sergey Bartunov, Matthew Botvinick, Daan Wierstra, and Timothy
  Lillicrap.
\newblock Meta-learning with memory-augmented neural networks.
\newblock In \emph{International conference on machine learning}, pp.\
  1842--1850, 2016.

\bibitem[Schulman et~al.(2015)Schulman, Levine, Abbeel, Jordan, and
  Moritz]{schulman2015trust}
John Schulman, Sergey Levine, Pieter Abbeel, Michael Jordan, and Philipp
  Moritz.
\newblock Trust region policy optimization.
\newblock In \emph{International Conference on Machine Learning}, pp.\
  1889--1897, 2015.

\bibitem[Smith \& Slone(2017)Smith and Slone]{smith2017developmental}
Linda~B Smith and Lauren~K Slone.
\newblock A developmental approach to machine learning?
\newblock \emph{Frontiers in psychology}, 8:\penalty0 2124, 2017.

\bibitem[Snell et~al.(2017)Snell, Swersky, and Zemel]{snell2017prototypical}
Jake Snell, Kevin Swersky, and Richard Zemel.
\newblock Prototypical networks for few-shot learning.
\newblock In \emph{Advances in Neural Information Processing Systems}, pp.\
  4080--4090, 2017.

\bibitem[Tenenbaum(1999)]{tenenbaum1999bayesian}
Joshua~Brett Tenenbaum.
\newblock \emph{A Bayesian framework for concept learning}.
\newblock PhD thesis, Massachusetts Institute of Technology, 1999.

\bibitem[Todorov et~al.(2012)Todorov, Erez, and Tassa]{todorov2012mujoco}
Emanuel Todorov, Tom Erez, and Yuval Tassa.
\newblock Mujoco: A physics engine for model-based control.
\newblock In \emph{Intelligent Robots and Systems (IROS), 2012 IEEE/RSJ
  International Conference on}, pp.\  5026--5033. IEEE, 2012.

\bibitem[Vinyals et~al.(2016)Vinyals, Blundell, Lillicrap, Wierstra,
  et~al.]{vinyals16matching}
Oriol Vinyals, Charles Blundell, Tim Lillicrap, Daan Wierstra, et~al.
\newblock Matching networks for one shot learning.
\newblock In \emph{Advances in Neural Information Processing Systems}, pp.\
  3630--3638, 2016.

\bibitem[Welling \& Teh(2011)Welling and Teh]{welling2011bayesian}
Max Welling and Yee~W Teh.
\newblock Bayesian learning via stochastic gradient langevin dynamics.
\newblock In \emph{Proceedings of the 28th International Conference on Machine
  Learning (ICML-11)}, pp.\  681--688, 2011.

\bibitem[Williams(1992)]{williams1992simple}
Ronald~J Williams.
\newblock Simple statistical gradient-following algorithms for connectionist
  reinforcement learning.
\newblock In \emph{Reinforcement Learning}, pp.\  5--32. Springer, 1992.

\end{thebibliography}
\bibliographystyle{iclr_2017}

\newpage
\begin{appendices}
\section{Supervised Learning}
\subsection{Regression}

\begin{figure}[h]
\makebox[\textwidth][c]{
\begin{adjustbox}{minipage=\linewidth,scale=1.2}
    \centering
    \begin{subfigure}[b]{0.325\textwidth}
        \centering
        \caption{MAML (M=1)}
        \makebox[\textwidth][c]{
        \includegraphics[width=1.1\textwidth]{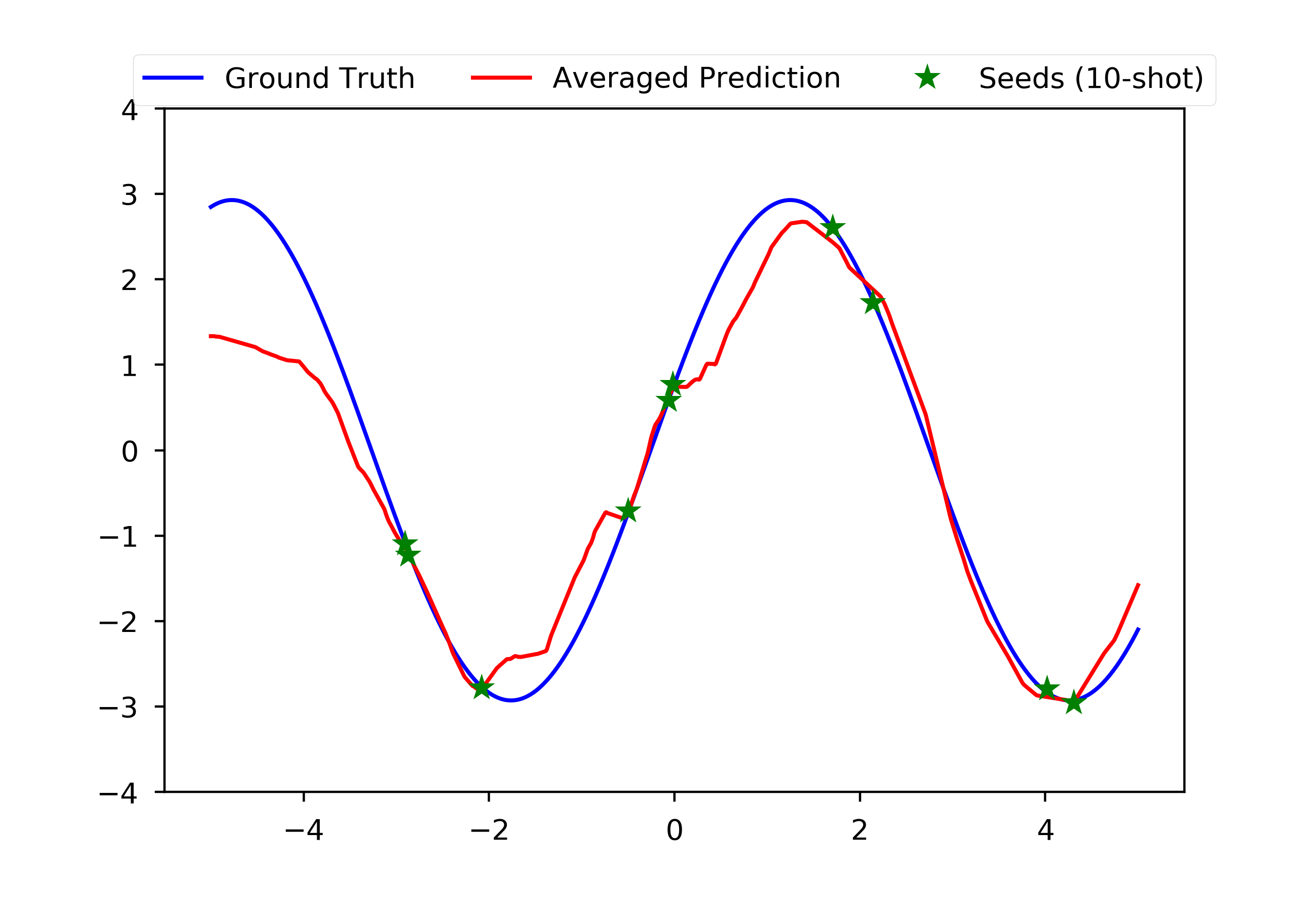}}
    \end{subfigure}
    \hfill
    \begin{subfigure}[b]{0.325\textwidth}
        \centering
        \caption{EMAML (M=20)}
        \makebox[\textwidth][c]{
        \includegraphics[width=1.1\textwidth]{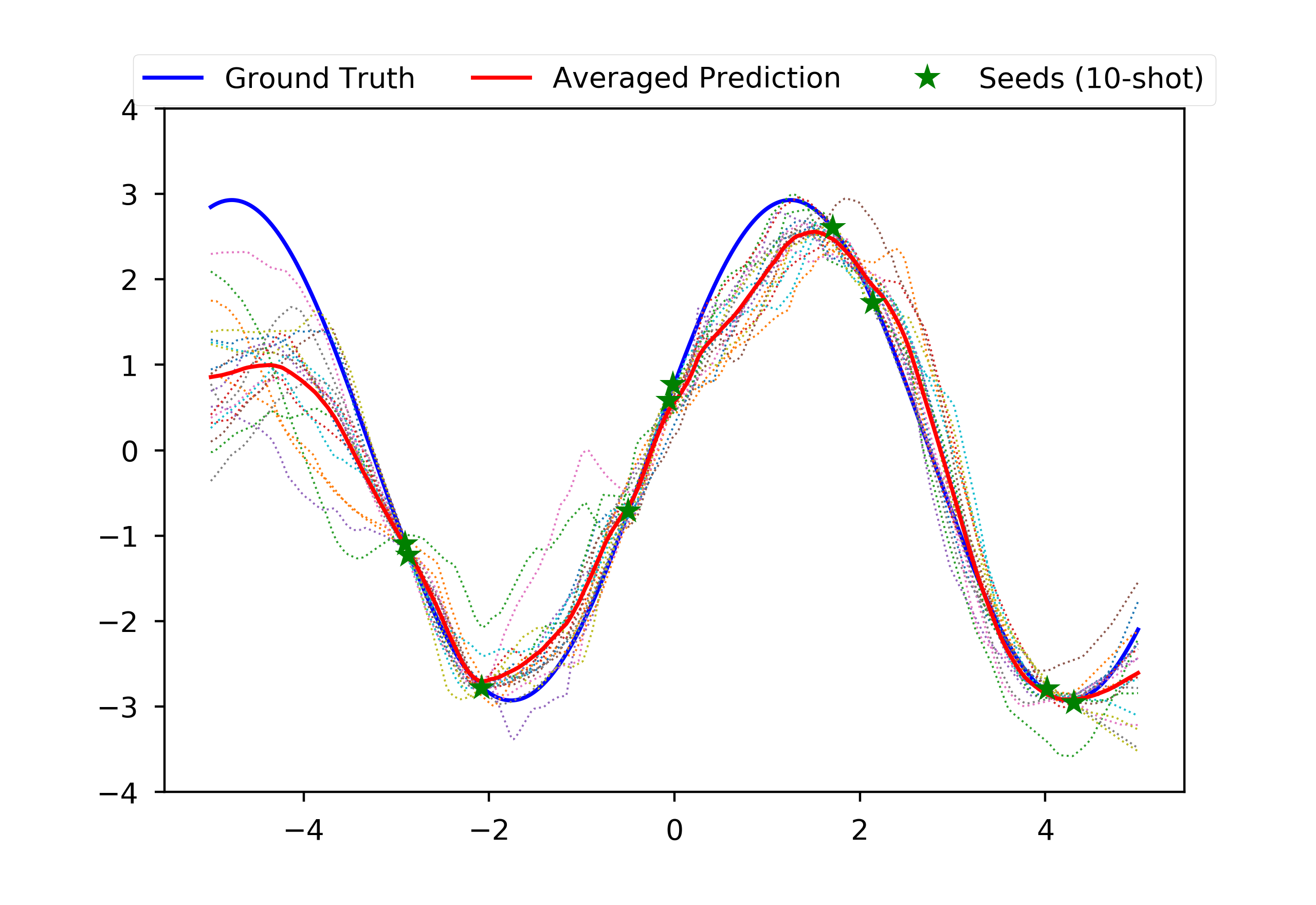}}
    \end{subfigure}
    \hfill
    \begin{subfigure}[b]{0.325\textwidth}
        \centering
        \caption{BMAML (M=20)}
        \makebox[\textwidth][c]{
        \includegraphics[width=1.1\textwidth]{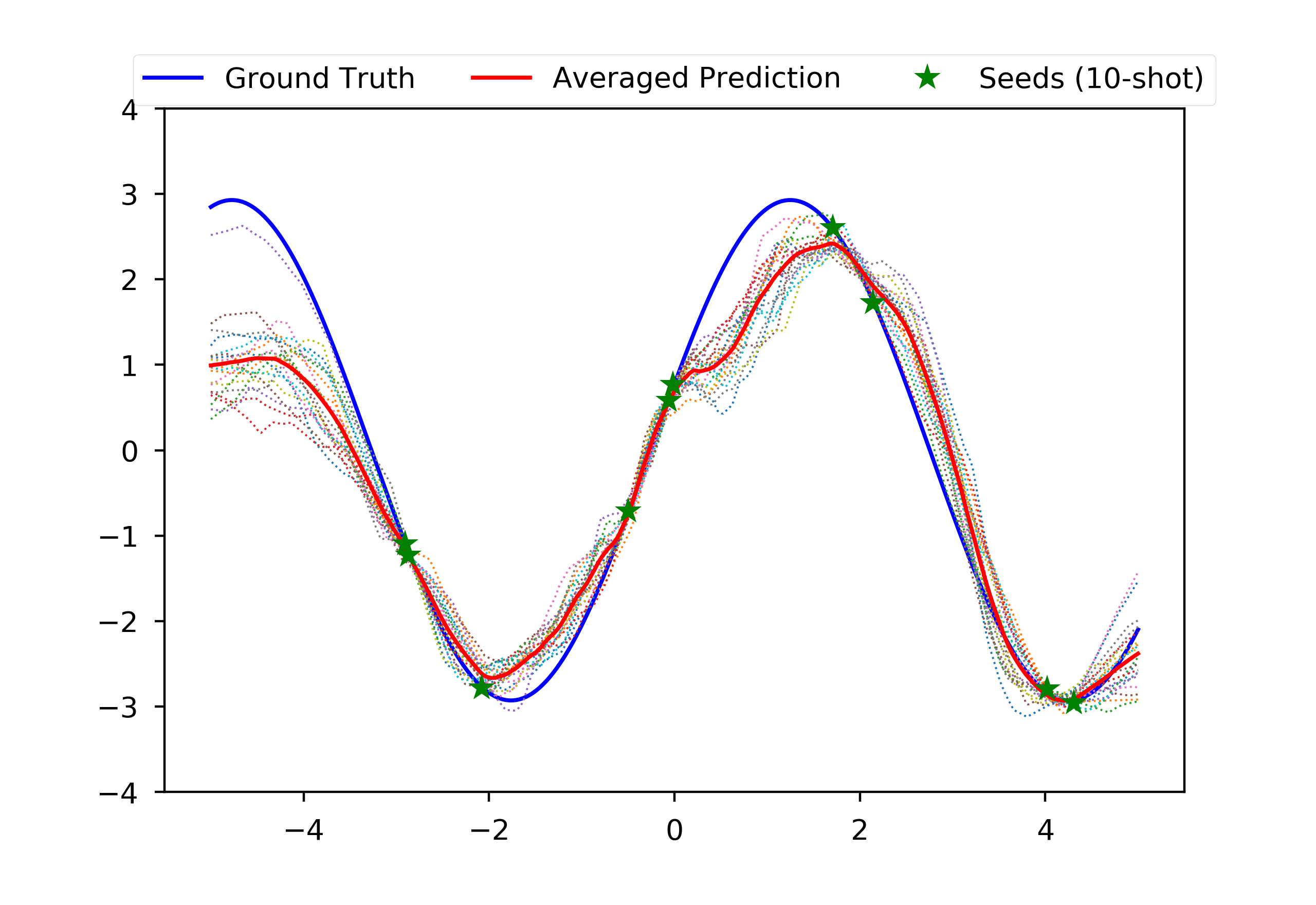}}
    \end{subfigure}

    \begin{subfigure}[b]{0.325\textwidth}
        \centering
        \makebox[\textwidth][c]{
        \includegraphics[width=1.1\textwidth]{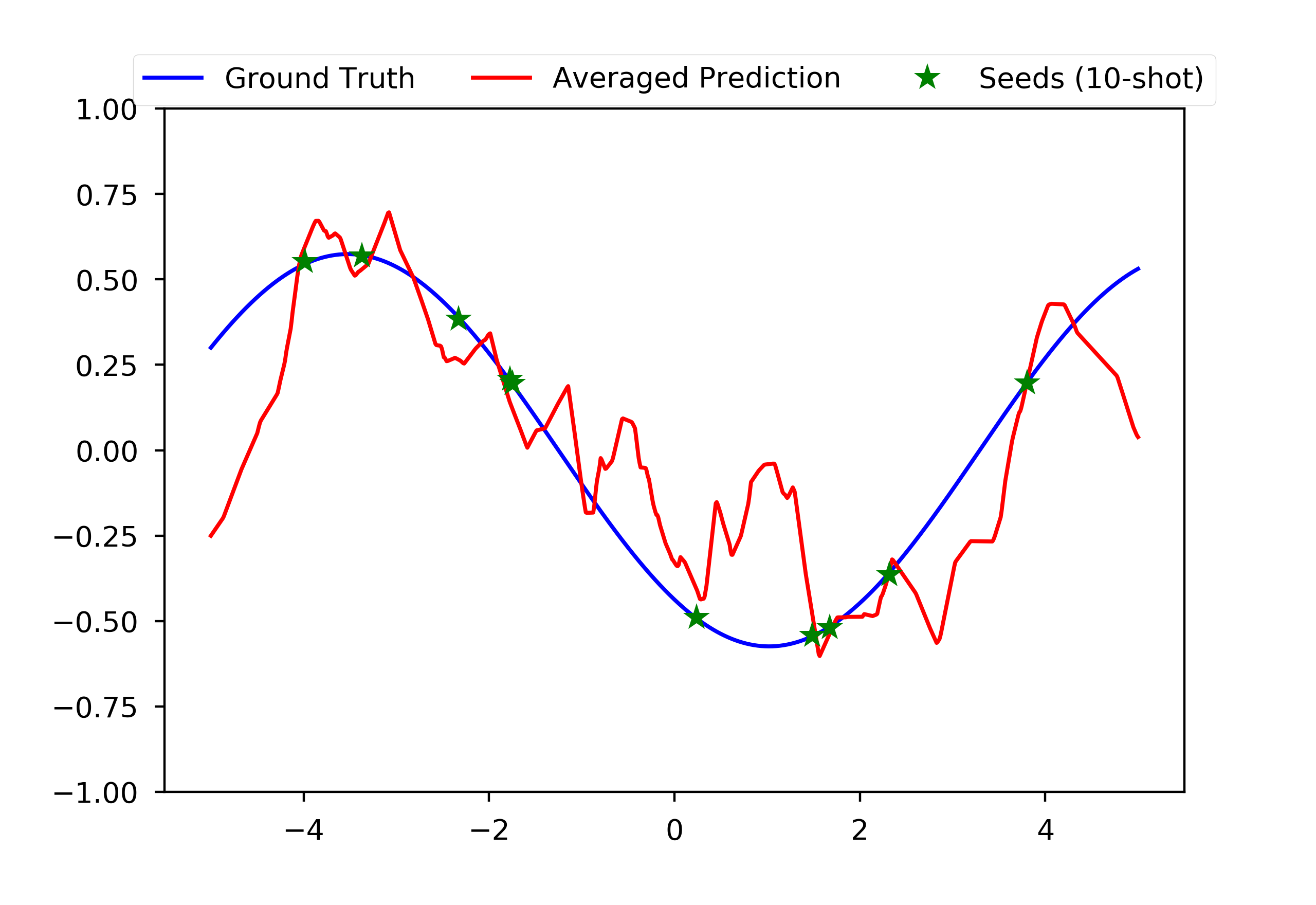}}
    \end{subfigure}
    \hfill
    \begin{subfigure}[b]{0.325\textwidth}
        \centering
        \makebox[\textwidth][c]{
        \includegraphics[width=1.1\textwidth]{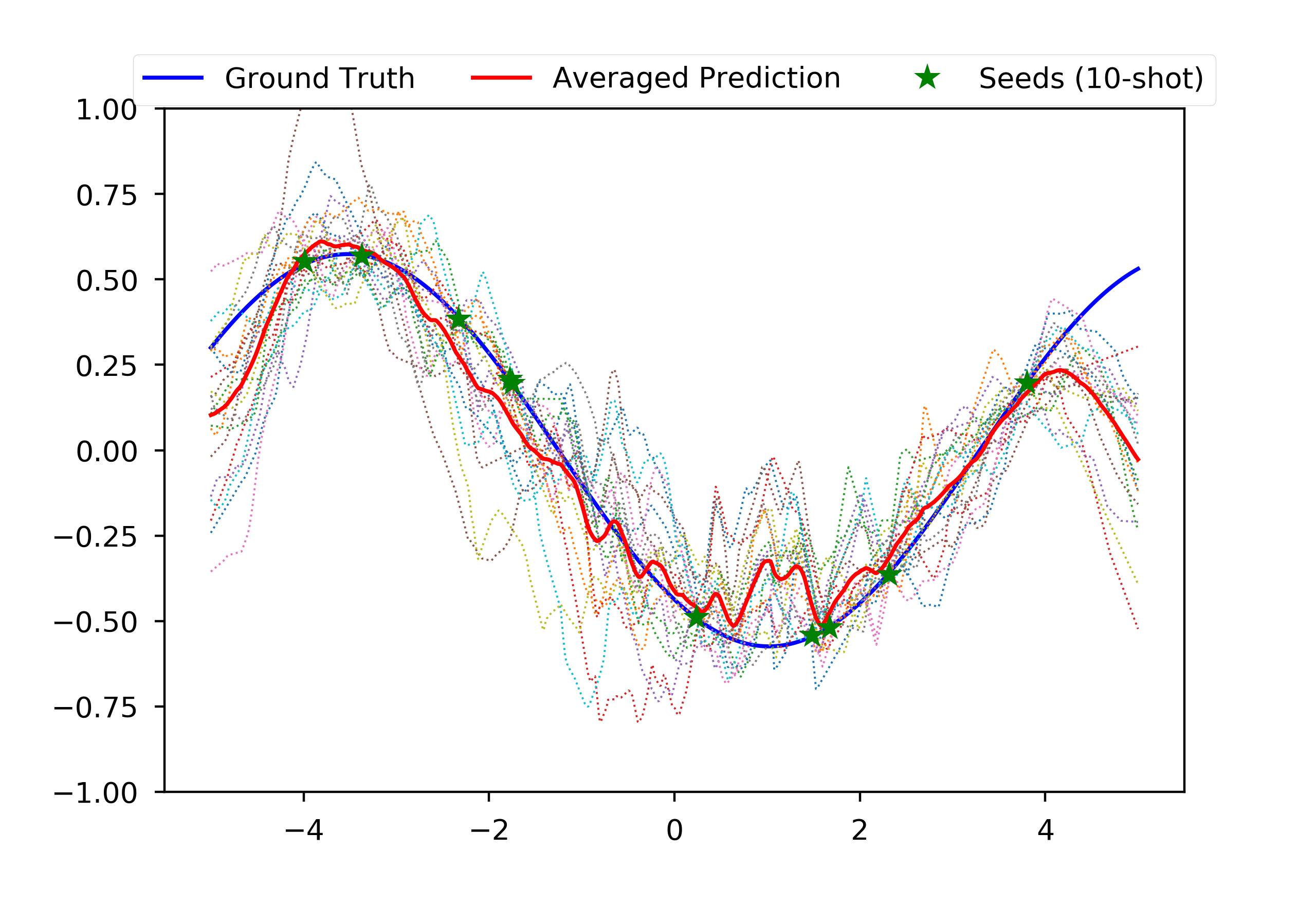}}
    \end{subfigure}
    \hfill
    \begin{subfigure}[b]{0.325\textwidth}
        \centering
        \makebox[\textwidth][c]{
        \includegraphics[width=1.1\textwidth]{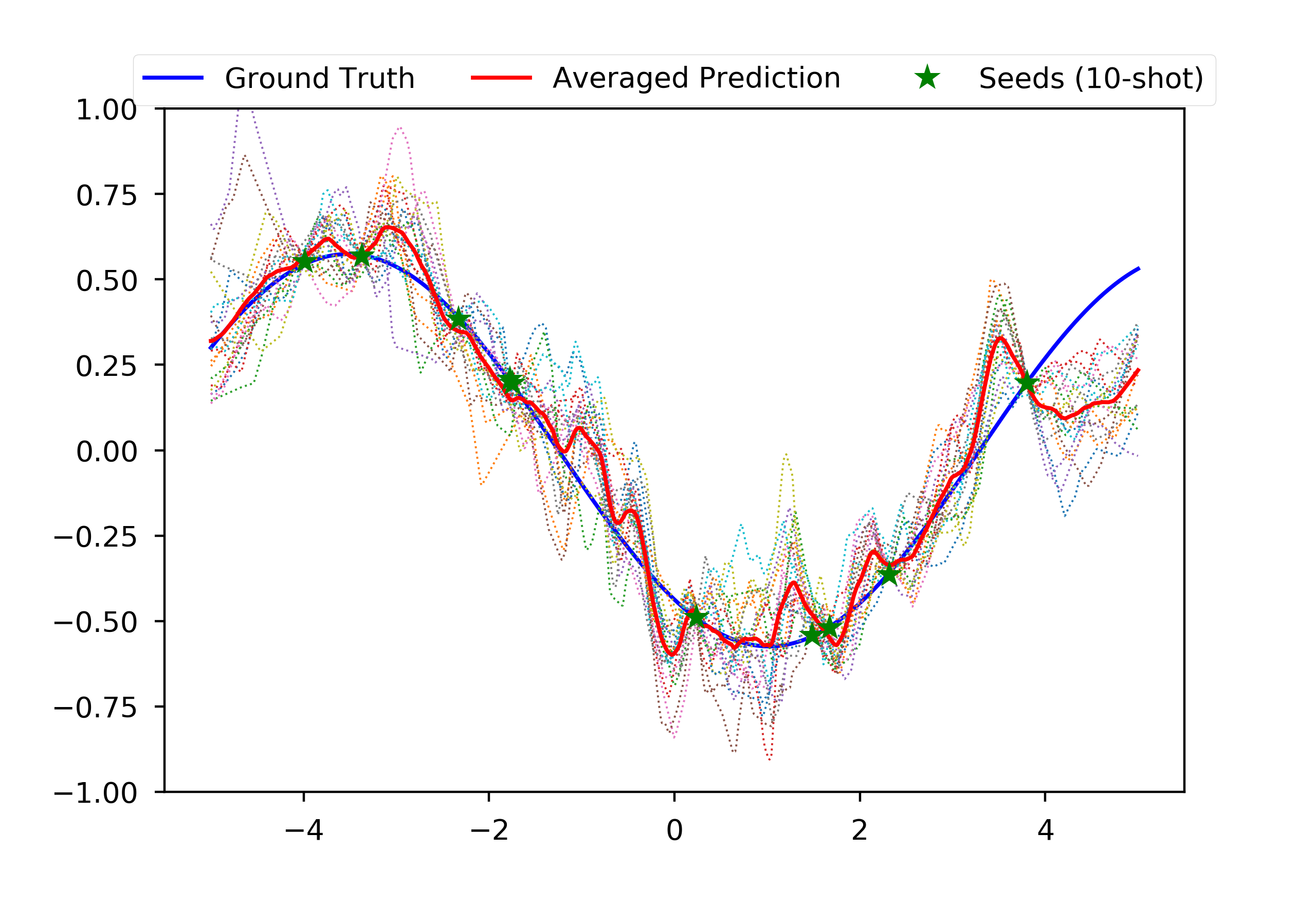}}
    \end{subfigure} 

    \begin{subfigure}[b]{0.325\textwidth}
        \centering
        \makebox[\textwidth][c]{
        \includegraphics[width=1.1\textwidth]{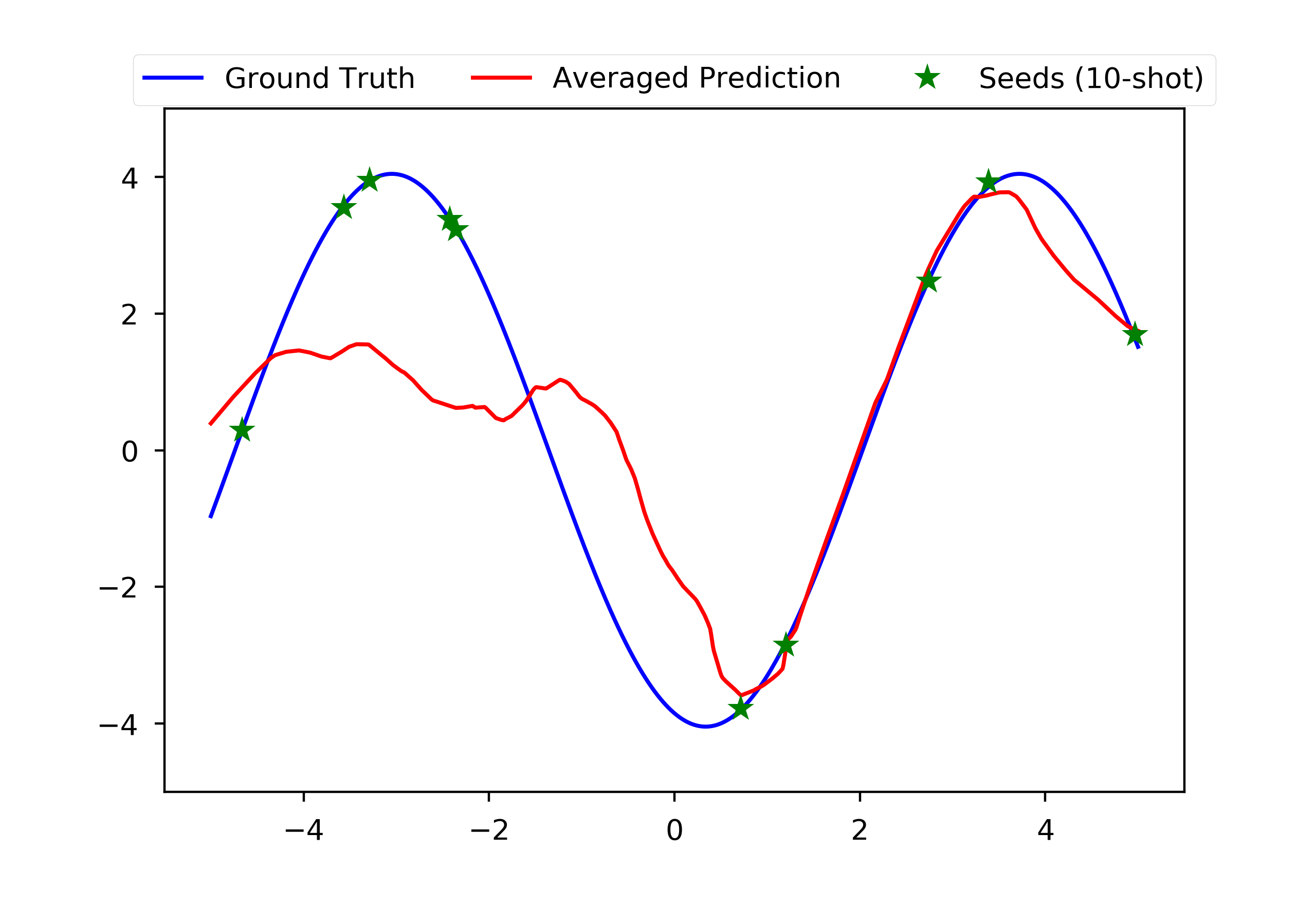}}
    \end{subfigure}
    \hfill
    \begin{subfigure}[b]{0.325\textwidth}
        \centering
        \makebox[\textwidth][c]{
        \includegraphics[width=1.1\textwidth]{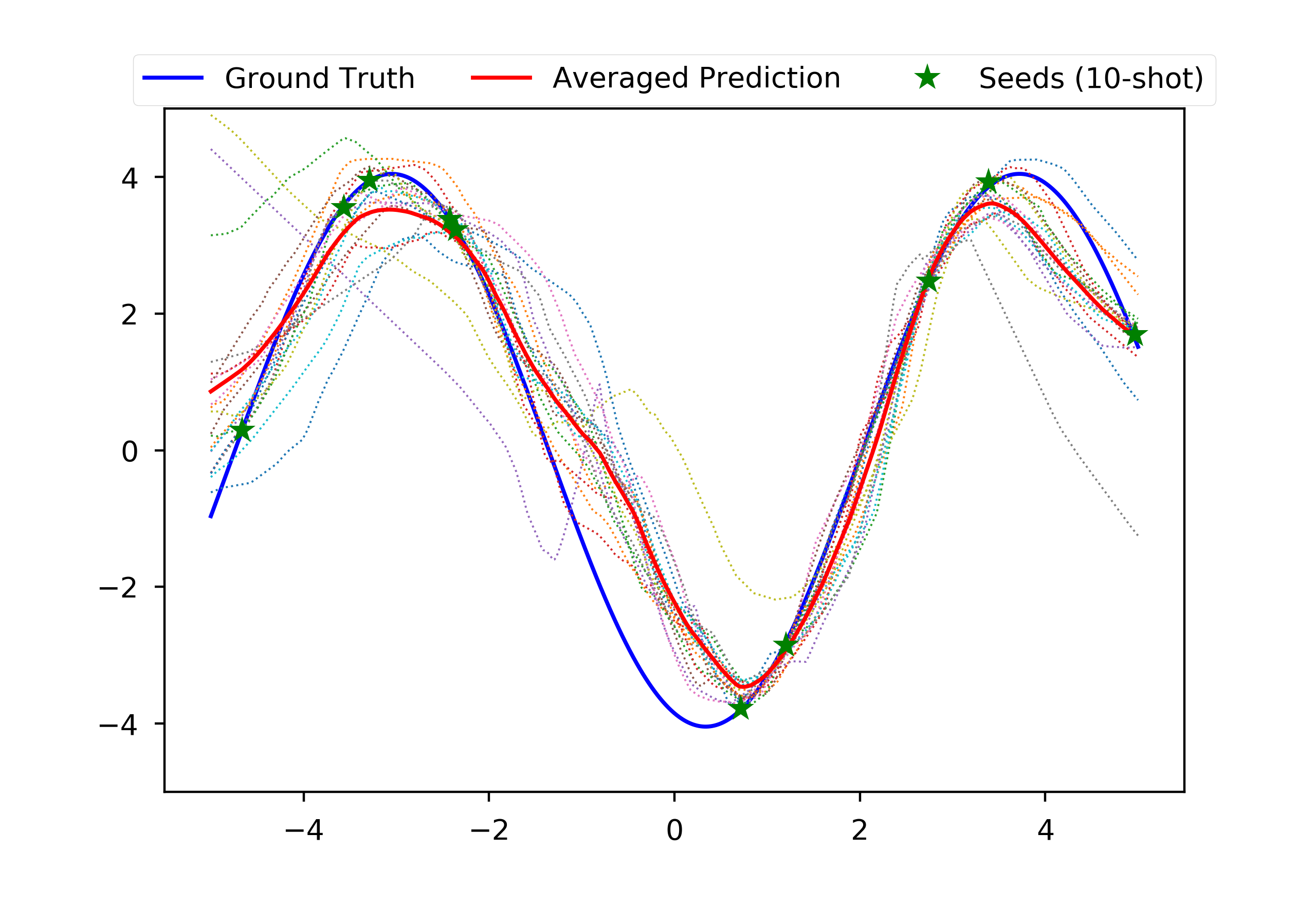}}
    \end{subfigure}
    \hfill
    \begin{subfigure}[b]{0.325\textwidth}
        \centering
        \makebox[\textwidth][c]{
        \includegraphics[width=1.1\textwidth]{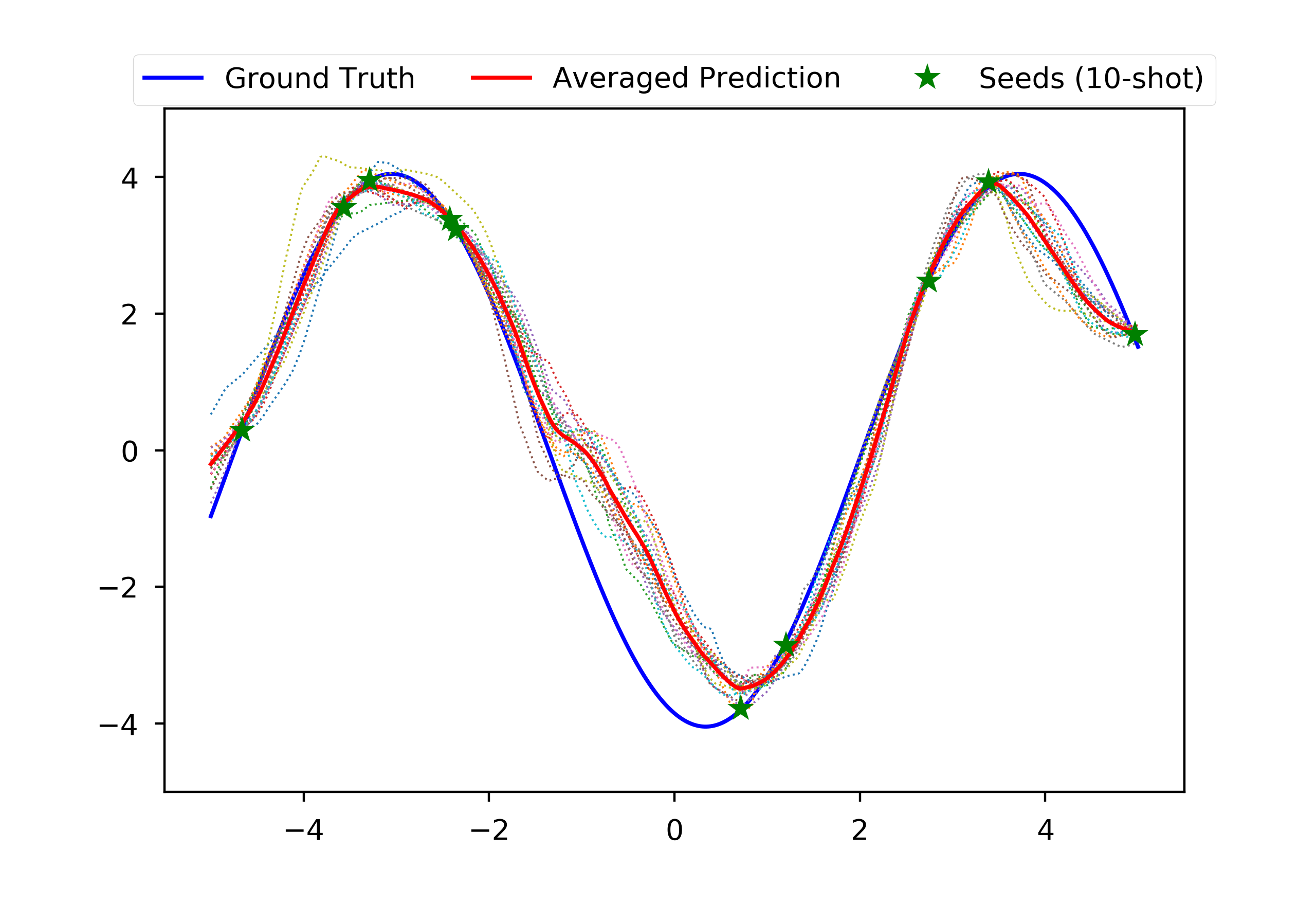}}
    \end{subfigure} 

    \begin{subfigure}[b]{0.325\textwidth}
        \centering
        \makebox[\textwidth][c]{
        \includegraphics[width=1.1\textwidth]{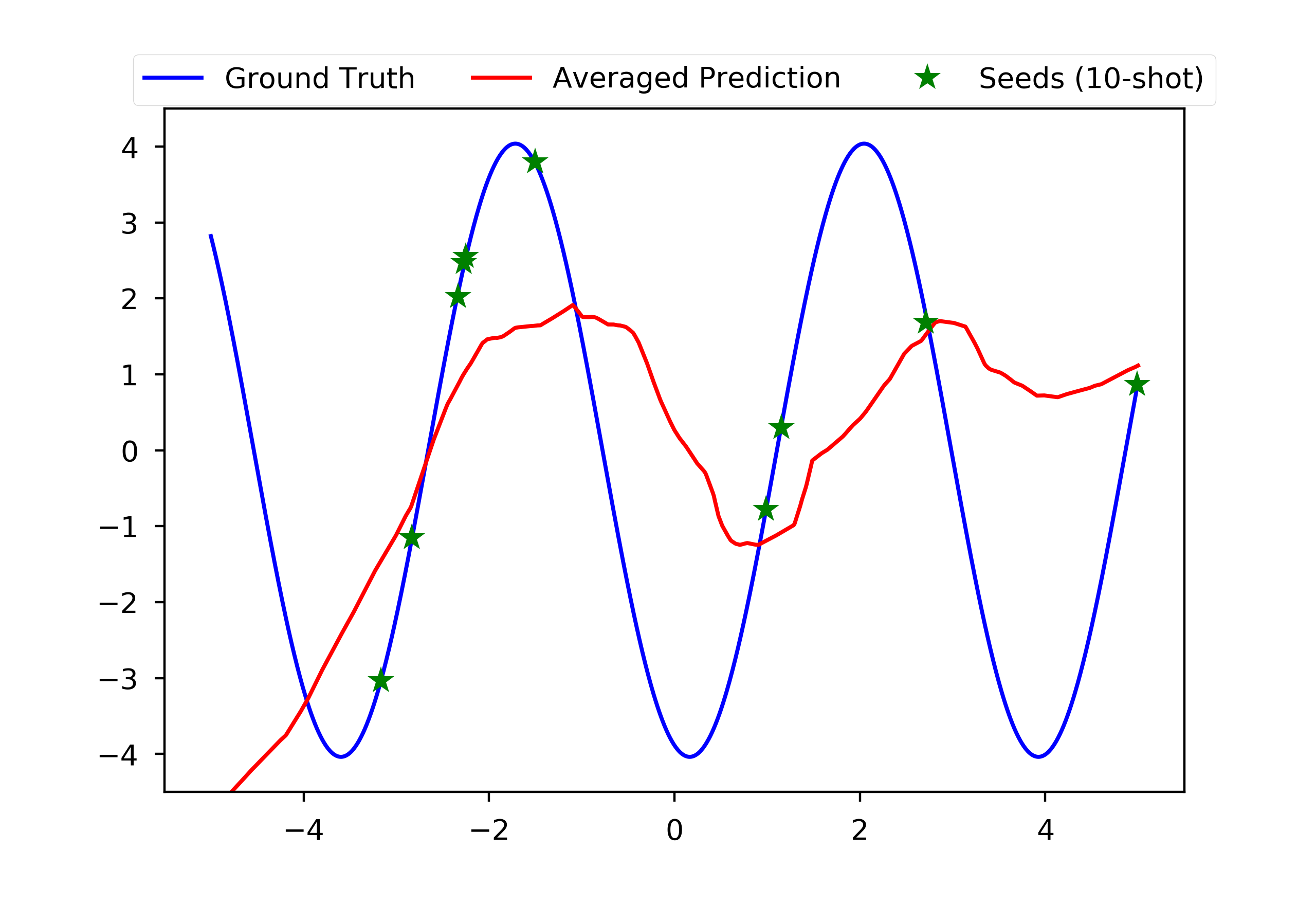}}
    \end{subfigure}
    \hfill
    \begin{subfigure}[b]{0.325\textwidth}
        \centering
        \makebox[\textwidth][c]{
        \includegraphics[width=1.1\textwidth]{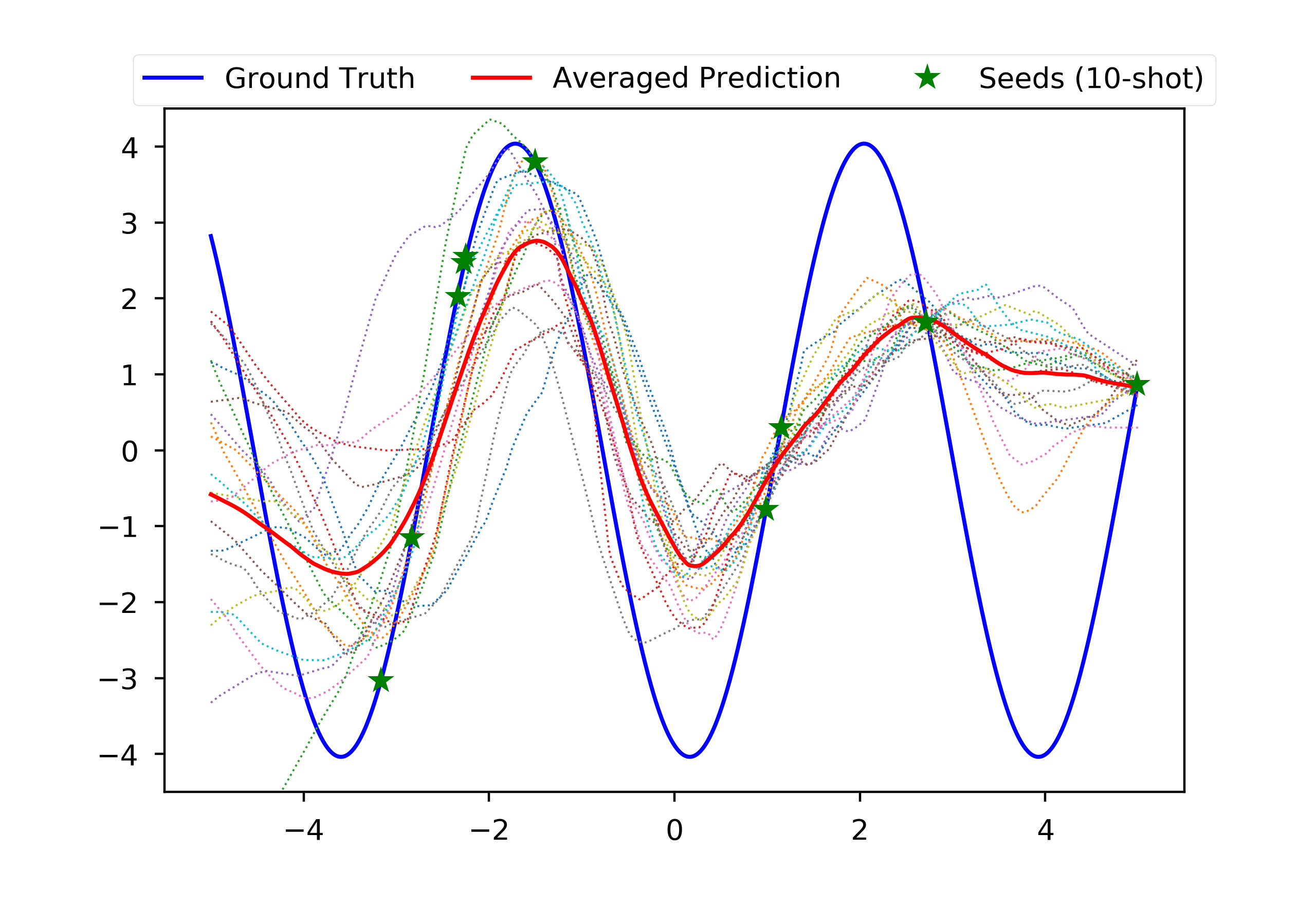}}
    \end{subfigure}
    \hfill
    \begin{subfigure}[b]{0.325\textwidth}
        \centering
        \makebox[\textwidth][c]{
        \includegraphics[width=1.1\textwidth]{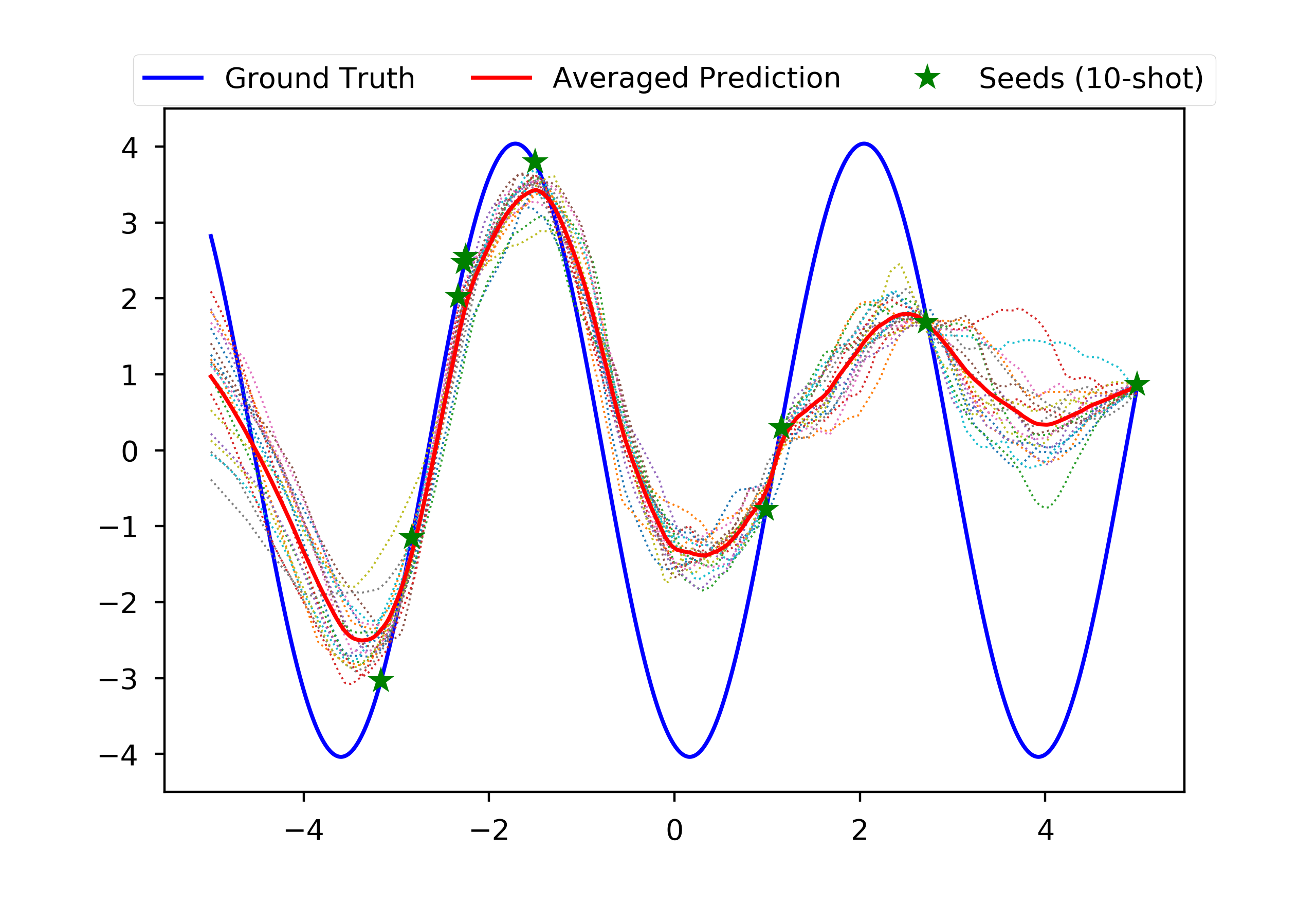}}
    \end{subfigure} 

    \begin{subfigure}[b]{0.325\textwidth}
        \centering
        \makebox[\textwidth][c]{
        \includegraphics[width=1.1\textwidth]{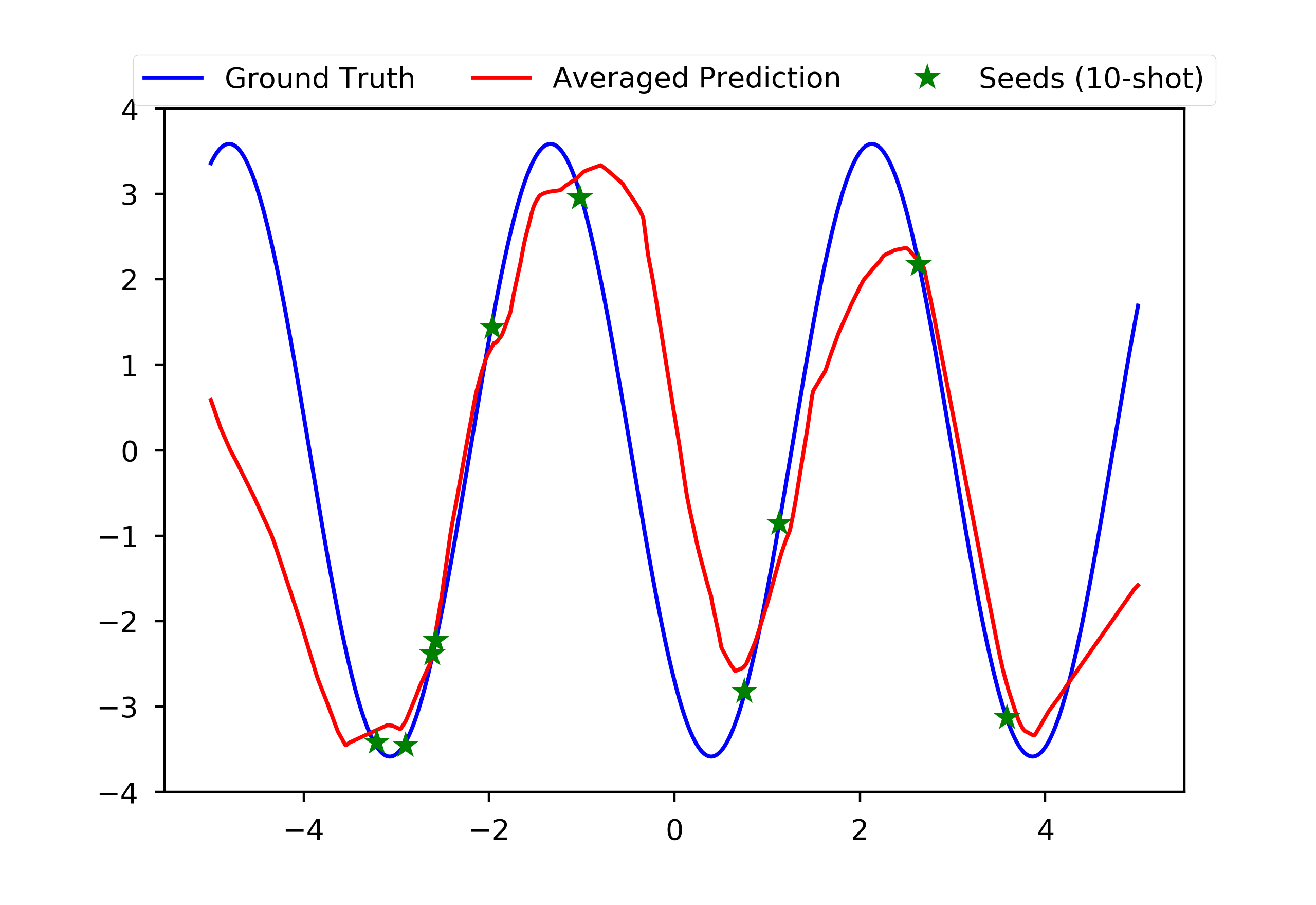}}
    \end{subfigure}
    \hfill
    \begin{subfigure}[b]{0.325\textwidth}
        \centering
        \makebox[\textwidth][c]{
        \includegraphics[width=1.1\textwidth]{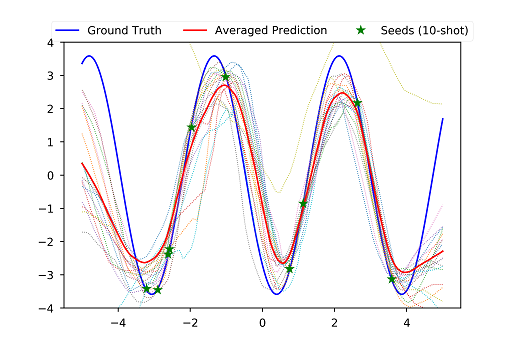}}
    \end{subfigure}
    \hfill
    \begin{subfigure}[b]{0.325\textwidth}
        \centering
        \makebox[\textwidth][c]{
        \includegraphics[width=1.1\textwidth]{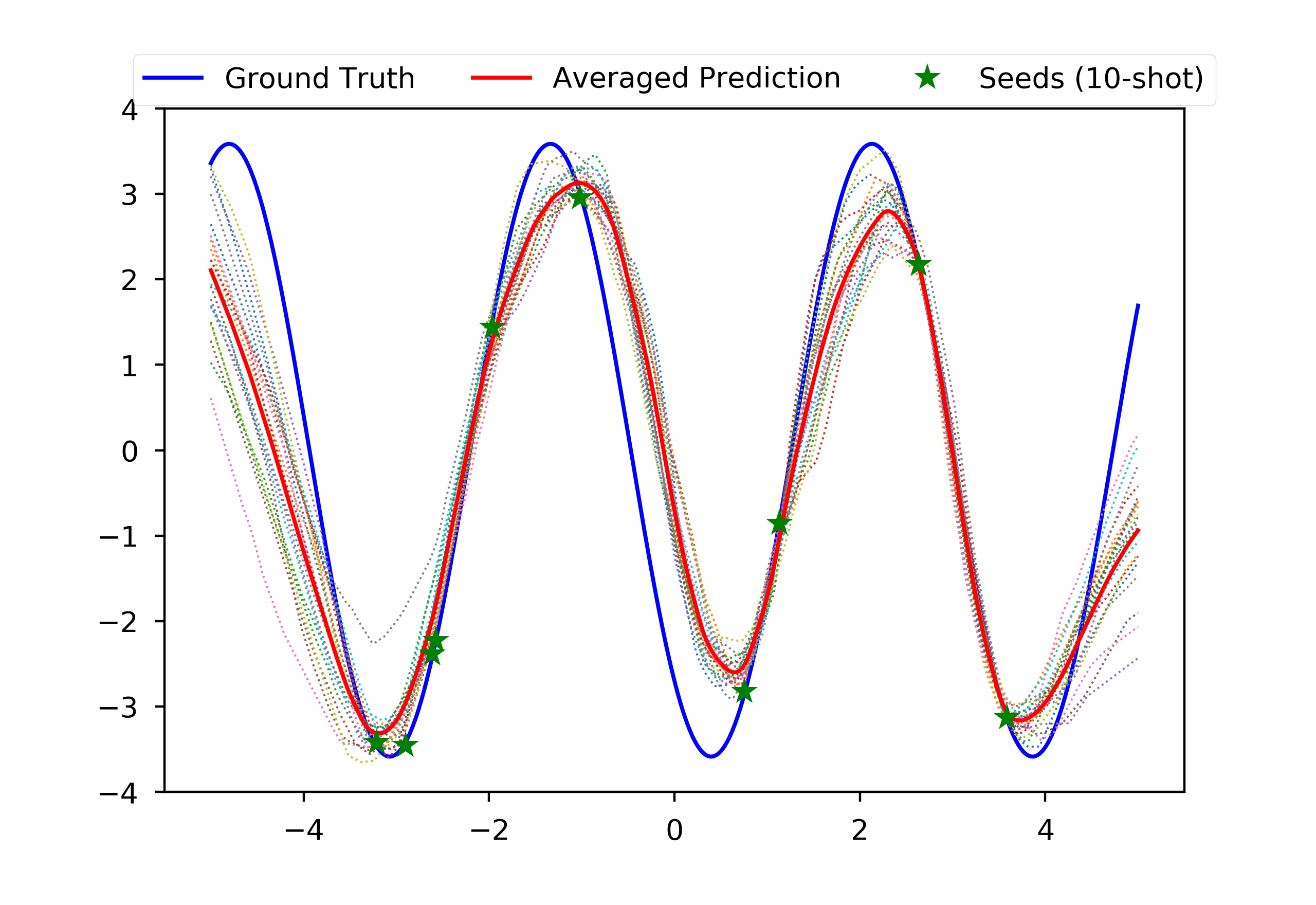}}
    \end{subfigure}
\end{adjustbox}    
}
\caption{Regression qualitative examples: randomly sampled tasks with 10 examples (10-shot) and 10 gradient updates for adaptation}
\label{fig:reg_samples}
\end{figure}

For regression, we used 10 tasks for each meta-batch and the meta-validation dataset $\cD_{\tau}^{\text{val}}$ is set to have the same size of the meta-training dataset $\cD_{\tau}^{\text{trn}}$~$(|\cD_{\tau}^{\text{trn}}| = |\cD_{\tau}^{\text{val}}| = K)$. During training, the number of steps $n$ for chaser is set to $n=1$ and also the number of steps $s$ for leader is set to $s=1$. We used different step sizes $\alpha$ for computing chaser and leader, 0.01 and 0.001, respectively.  This allows the leader to stay nearby the chaser but toward the target posterior and stabilized the training. The models were trained with using different size of training dataset $|\cT|$, the number of tasks observable during training, and we trained the model over 10000 epochs for $|\cT|=100$ and 1000 epochs for $|\cT|=1000$. In Fig.~\ref{fig:reg_samples}, we show the qualitative results on randomly sampled sinusoid task and we used 5 update steps. The task-train posterior $p(\theta_{\tau}|\cD_{\tau}^{\text{trn}})$ decomposes into the train data likelihood and parameter prior as $p(\theta_{\tau}|\cD_{\tau}^{\text{trn}}) \propto p(\cD_{\tau}^{\text{trn}}|\theta_{\tau})p(\theta_{\tau})$ and this is formulated as:
$$
p(\theta_{\tau}|\cD_{\tau}^{\text{trn}}) \propto \prod_{(x, y) \in \cD_{\tau}^{\text{trn}}}{\cN(y|f_{W}(x), \gamma^{-1})} \prod_{w \in W}{\cN(w|0, \lambda^{-1})} \textrm{Gamma}(\gamma| a, b) \textrm{Gamma}(\lambda| a', b') 
$$
where $\theta_{\tau}$ consists of network parameters $W$ and scaling parameters $\gamma, \lambda$. In all experiments, we set Gamma distribution hyper-parameters as $a=2.0, b=0.2$ and $a'=2.0, b'=2.0$. During meta-update with chaser-loss, we used Adam optimizer \citep{kingma2014adam} with learning rate $\beta=0.001$.

\subsection{Classification}
\label{appendix:class}
All models and experiments on the \textit{mini}Imagenet classification task are trained with using the same network architecture and 16 tasks are used for each meta-batch during training. Each task is defined by randomly selected 5 classes with one instance of each class to adapt the model and it is evaluated on unseen instances within the selected 5 classes. We used the meta-validation dataset $\cD_{\tau}^{\text{val}}$ containing one example per each class for the 5-way 1-shot setting. This reduced the computational cost and also improved the performance of all models. During training, the number of steps for chaser and leader both are set to 1~$(n=s=1)$. The chaser and leader used step size $\alpha=0.01$ and $\alpha=0.005$, respectively. The meta-update was done by using Adam optimizer ($\beta=0.0005$). The models were trained with using different size of training dataset $|\cT|$ and we trained the model with $|\cT|=800000$ and 1 epoch. With $|\cT|=10000$, the model was trained over 40 epochs. The task-train posterior $p(\theta_{\tau}|\cD_{\tau}^{\text{trn}})$ for classification is slightly different to the regression task due to using softmax for the data likelihood.
$$
p(\theta_{\tau}|\cD_{\tau}^{\text{trn}}) \propto \prod_{(x, y) \in \cD_{\tau}^{\text{trn}}}{p(y|f_{W}(x))} \prod_{w \in W}{\cN(w|0, \lambda^{-1})} \textrm{Gamma}(\lambda| a, b) 
$$ 
where the hyper-parameters for Gamma distribution were set as $a=2.0, b=0.2$ or $a=1.0, b=0.1$ in our experiments.

\section{Active Learning}
\begin{algorithm}[h]
\begin{algorithmic}[1]
\caption{Active Learning on Image Classification}
\STATE Sample a few-shot labeled dataset $\cD_{\tau}$ and a pool of unlabeled dataset $\cX_{\tau}$ of task $\tau$
\STATE Initialize $\Ta_\tau \law \Ta_0^{*}$
\STATE Update $\Ta_\tau \law \textrm{SVGD}_n(\Ta_\tau;\cD_\tau, \alpha)$  
\WHILE{$\cX_{\tau}$ is not empty}
    \STATE Select $x' \law \argmax_{x \in \cX_\tau} \eH[y|x, \Ta_\tau]$ and remove $x'$ from $\cX_\tau$
    \STATE Request $y'$ of $x'$
    \STATE Update $\cD_\tau \law \cD_\tau \cup \{(x', y')\}$
    \STATE Update $\Ta_\tau \law \textrm{SVGD}_n(\Ta_\tau;\cD_\tau, \alpha)$  
\ENDWHILE
\label{algo:active}
\end{algorithmic}
\end{algorithm}

\section{Reinforcement Learning}

\subsection{Locomotion} \label{rl_locomotion}

The locomotion experiments require two simulated robots, a planar cheetah and 3D quadruped ones (called as ant), and two individual goals, to run in a particular direction or at a particular velocity.
For the ant goal velocity, a positive bonus reward at each timestep is added to prevent the ant from ending the episode.
In those experiments, the timestep in each episodes is 200, the number of episode per each inner update, $K$ is 10 except the ant goal direction task, in which 40 episodes for each inner update is used, because of task complexity.
The number of tasks per each meta update is 20, and the models are trained for up to 200 meta iterations.

\subsection{Used Methods} \label{rl_methods}
We evaluate our proposed method on two cases, SVPG-TRPO vs VPG-TRPO and SVPG-Chaser vs VPG-Reptile. We describe the methods in this subsection, except VPG-TRPO, because this is MAML when $M=1$. 

\subsubsection{SVPG-TRPO}
This method is to use SVPG as inner update and TRPO as meta update, which is following to a simple Bayesian meta-learning manner.
In $K$-shot reinforcement learning on this method, $K$ episodes from each policy particles and task $\tau$ (total number of episode is $KM$), and the corresponding rewards are used for task learning on the task.
This method gets the above data ($\cD^{trn}_\tau$) from $\Ta_0$, and updates the parameters $\Ta_0$ to $\Ta_\tau^n$ with $\cD^{trn}_\tau$ and SVPG. 
After getting few-shot learned parameters ($\Ta_\tau^n$), our method get new data ($\cD^{val}_\tau$) from $\Ta_\tau^n$.
After all the materials for meta learning have been collected, our method finds the meta loss with few-shot learned particles and task-validation set, $\cD^{val}_\tau$. On meta-learning, TRPO \citep{schulman2015trust} is used as MAML \citep{maml} for validating inner Bayesian learning performance.
The overall algorithm is described in Algorithm \ref{algo:bml_rl}.

\begin{algorithm}[h]
\begin{algorithmic}[1]
\caption{Simple Bayesian Meta-Learning for Reinforcement Learning}
\STATE Initialize $\Ta_0$ 
\FOR{$t=0,\dots$ until converge}
\STATE Sample a mini-batch of tasks $\cT_t$ from $p(\cT)$
    \FOR {each task $\tau \in \cT_t$}
    \STATE Sample trajectories $\cD^{\text{trn}}_\tau$ with $\Ta_0$ in $\tau$
    \STATE Compute chaser $\Ta_\tau^{n} = \textrm{SVPG}(\Ta_0;\cD_\tau^{\text{trn}})$
    \STATE Sample trajectories $\cD^{\text{val}}_\tau$ with $\Ta_{\tau}^{n}$ in $\tau$
    \ENDFOR
\STATE $\Ta_0 \law \Ta_0 - \bt \grad_{\Ta_0}  \sum_{\tau \in \cT_t}{\cL_{\tau}^{meta}(\Ta_{\tau}^{n};\cD_{\tau}^{\text{val}})}$    
\ENDFOR
\label{algo:bml_rl}
\end{algorithmic}
\end{algorithm}

\subsubsection{SVPG-Chaser}
This method is to use SVPG as inner-update and chaser loss for meta-update to maintain uncertainty.
Different to supervised learning, this method updates leader particles just with $\cD_\tau^{\text{val}}$ in policy gradient update manner.
Same chaser loss to supervised learning ones is consistently applied to evaluating the chaser loss extensibility. 
The chaser loss in RL changes the meta update from a policy gradient problem to a problem similar to imitation learning.
Unlike conventional imitation learning with given expert agent, this method keeps the uncertainty provided by the SVPG by following one more updated agent, and ultimately ensures that the chaser agent is close to the expert.
Compared to Algorithm \ref{algo:bml_rl}, this method adds updating the leader and changes the method of meta update like Algorithm \ref{algo:bml_rl_chaser}.

\begin{algorithm}[h]
\begin{algorithmic}[1]
\caption{Bayesian Meta-Learning for Reinforcement Learning with Chaser Loss}
\STATE Initialize $\Ta_0$ 
\FOR{$t=0,\dots$ until converge}
\STATE Sample a mini-batch of tasks $\cT_t$ from $p(\cT)$
    \FOR {each task $\tau \in \cT_t$}
    \STATE Sample trajectories $\cD^{\text{trn}}_\tau$ with $\Ta_0$ in $\tau$
    \STATE Compute chaser $\Ta_\tau^n = \textrm{SVPG}(\Ta_0;\cD_\tau^{\text{trn}})$
    \STATE Sample trajectories $\cD^{\text{val}}_\tau$ with $\Ta_{\tau}^n$ in $\tau$
    \STATE Compute leader $\Ta_\tau^{n+s} = \textrm{SVPG}(\Ta_\tau^n;\cD_\tau^{\text{val}})$
    \ENDFOR
\STATE $\Ta_0 \law \Ta_0 - \bt \grad_{\Ta_0} \sum_{\tau\in\cT_t}D(\Ta_\tau^n||\textrm{stopgrad}(\Ta_\tau^{n+s}))$
\ENDFOR
\label{algo:bml_rl_chaser}
\end{algorithmic}
\end{algorithm}

\subsubsection{VPG-Reptile}

Reptile \citep{reptile} solved meta learning problem by using only $1^\text{st}$-order derivatives, and used meta update in parameter space.
We design a version of meta loss similar to Reptile to verify the performance of chaser loss in RL problem.
This method computes the chaser $\Ta_\tau^n$ using $\cD_\tau^{\text{trn}}$ and then calculates the euclidean distance between this parameter and the global parameter as a meta loss (to prevent the gradient from being calculated through the chaser parameter to maintain the $1^\text{st}$-order derivatives).
The overall algorithm is described in Algorithm \ref{algo:bml_rl_reptile}.

\begin{algorithm}[h]
\begin{algorithmic}[1]
\caption{VPG-Reptile}
\STATE Initialize $\Ta_0$ 
\FOR{$t=0,\dots$ until converge}
\STATE Sample a mini-batch of tasks $\cT_t$ from $p(\cT)$
    \FOR {each task $\tau \in \cT_t$}
    \STATE Sample trajectories $\cD^{\text{trn}}_\tau$ with $\Ta_0$ in $\tau$
    \STATE Compute chaser $\Ta_\tau^n = \textrm{VPG}(\Ta_0;\cD_\tau^{\text{trn}})$
    \ENDFOR
\STATE $\Ta_0 \law \Ta_0 - \bt \grad_{\Ta_0} \sum_{\tau\in\cT_t}D(\Ta_0||\textrm{stopgrad}(\Ta_\tau^{n}))$
\ENDFOR
\label{algo:bml_rl_reptile}
\end{algorithmic}
\end{algorithm}

\subsection{Experimental Details} \label{rl_setting}
Inner update learning rate and the number of inner update are set as 0.1 and 1 for all experiments, which are locomotion (ant/cheetah goal velocity and ant/cheetah goal direction) and 2D-Navigation experiments.
Meta update learning rate is set as 0.1 for ant goal direction and 0.01 for other experiments.
$\eta$, the parameter that controls the strength of exploration in SVPG is set as 0.1 for ant velocity experiment with SVPG-Chaser, ant goal direction and 2D Navigation with SVPG-Chaser, and 1.0 for other experiments.
Each plots are based on an mean and a standard deviation from three different random seed.
The subsumed results are plotted with the average reward of maximum task rewards in during of particles.

\subsection{Additional Experiment Results} \label{rl_additional_res}
\subsubsection{2D Navigation}
\begin{figure}[h]
\makebox[\textwidth][c]{
\begin{adjustbox}{minipage=\linewidth,scale=1.2}
    \centering
    \begin{subfigure}[b]{0.35\textwidth}
        \makebox[1.1\textwidth][c]{
        \includegraphics[width=\textwidth]{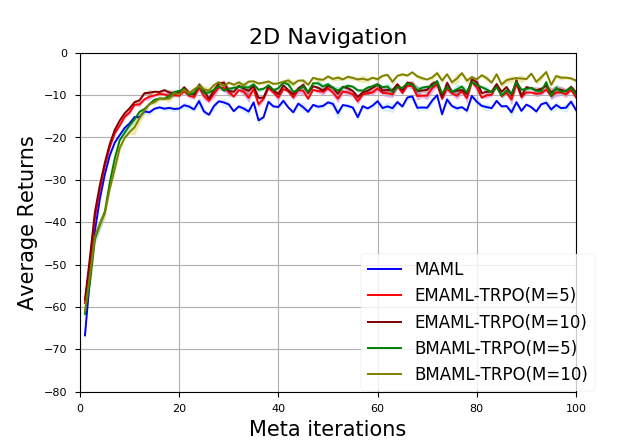}}
        \caption{}
    \end{subfigure}
    \begin{subfigure}[b]{0.35\textwidth}
        \makebox[1.1\textwidth][c]{
        \includegraphics[width=\textwidth]{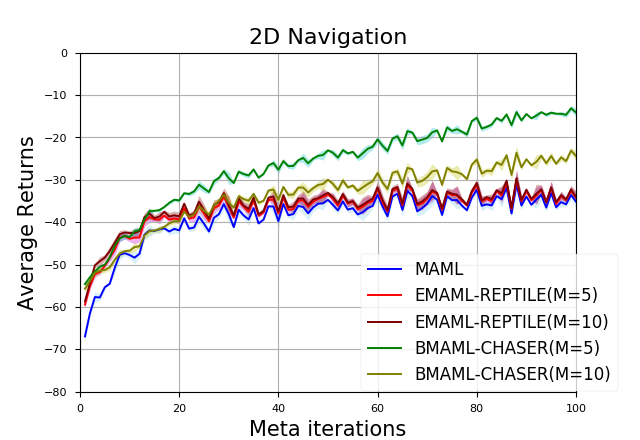}}
        \caption{}
    \end{subfigure}
\end{adjustbox}} 
    \caption{2D Navigation: (a) and (b) are results of SVPG-TRPO vs VPG-TRPO and SVPG-Chaser vs VPG-Reptile with three different random seed, respectively.}
    \label{fig:bmaml_rl_point_plot}
\end{figure}

We also compare the models on the toy experiment designed in previous work \citep{maml}, 2D Navigation. 
This experiment is a set of tasks where agent must move to different goal positions in 2D, which is randomly set for each task within a unit square.
The observation is the current 2D position, and actions correspond to velocity clipped to be in the range [-0.1, 0.1]. The reward is the negative squared distance between the goal and the current position, and episodes terminate when the agent is within 0.01 of the goal or at the timestep = 100.
We used 10 episodes per each inner update ($K=10$) and 20 tasks per each meta update.
The models are trained for up to 100 meta iterations.
The policy network has two hidden layers each with 100 ReLU units. We tested the number of particles for $M \in \{1, 5, 10\}$ with $M=1$ only for non-ensembled MAML.
Same to above locomotion experiments, we compare the models as SVPG-TRPO vs VPG-TRPO and SVPG-Chaser vs VPG-Reptile. As shown in Fig. \ref{fig:bmaml_rl_point_plot}, BMAML showed better performance than EMAML on both comparisons.

\end{appendices}
\end{document}